\pdfoutput=1

\documentclass[11pt]{article}

\usepackage[final]{acl}
\usepackage{float}
\usepackage{times}
\usepackage{latexsym}

\usepackage[T1]{fontenc}

\usepackage[utf8]{inputenc}

\usepackage{microtype}

\usepackage{inconsolata}

\usepackage{graphicx}

\usepackage{booktabs} 
\usepackage{longtable} 
\usepackage{xcolor}
\usepackage{ltablex}
\keepXColumns 

\title{Warm Up Before You Train: Unlocking General Reasoning in Resource-Constrained Settings}

\author{
  Safal Shrestha \quad Minwu Kim \quad Aadim Nepal \quad Anubhav Shrestha \quad Keith Ross\thanks{Corresponding author: \texttt{keithwross@nyu.edu}} \\
  \\
  Department of Computer Science, New York University Abu Dhabi\\
}

\begin{document}
\maketitle

\begin{abstract}
Designing effective reasoning-capable LLMs typically requires training using Reinforcement Learning with Verifiable Rewards (RLVR) or distillation with carefully curated Long Chain of Thoughts (CoT), both of which depend heavily on extensive training data. This creates a major challenge when the amount of quality training data is scarce. 
We propose a sample-efficient, two-stage training strategy to develop reasoning LLMs under limited supervision. In the first stage, we "warm up" the model by distilling Long CoTs from a toy domain, namely, Knights \& Knaves (K\&K) logic puzzles to acquire general reasoning skills. In the second stage, we apply RLVR to the warmed-up model using a limited set of target-domain examples. Our experiments demonstrate that this two-phase approach offers several benefits: $(i)$ the warmup phase alone facilitates generalized reasoning, leading to performance improvements across a range of tasks, including MATH, HumanEval\textsuperscript{+}, and MMLU-Pro; $(ii)$ When both the base model and the warmed-up model are RLVR trained on the same small dataset ($\leq100$ examples), 
the warmed-up model consistently outperforms the base model;
$(iii)$  Warming up before RLVR training allows a model to maintain cross-domain generalizability even after training on a specific domain; $(iv)$ Introducing warmup in the pipeline improves not only accuracy but also overall sample efficiency during RLVR training. The results in this paper highlight the promise of warmup for building robust reasoning LLMs in data-scarce environments.\footnote{Our code is available \href{https://github.com/safal312/warmup-before-you-train/}{here}.}\footnote{Our model weights and dataset are available \href{https://huggingface.co/collections/safal312/warmup-before-you-train-6829ddbc5e6f0c59f4a86a98}{here}.}
\end{abstract}

\section{Introduction}

Reasoning-capable large language models (LLMs) have driven a major shift in artificial intelligence, particularly in tasks requiring multi-step, cognitively complex problem solving \cite{guo2025deepseek,team2025kimi,openai_o1}. These models generate long CoTs, capturing reasoning behaviors such as self-reflection, self-correction, and hypothesis testing \cite{guo2025deepseek,team2025kimi,qin2024o1,huang2024o1,gao2024meta}. 
The CoTs mirror human-like problem-solving and have shown a notable improvement in accuracy on various benchmarks \cite{qin2024o1,huang2024o1,gao2024meta,xiang2025towards}.

Most prior approaches to training reasoning-capable models rely on either Reinforcement Learning with Verifiable Rewards (RLVR) or distillation with curated long-CoT demonstrations in specific domains such as math and coding \cite{guo2025deepseek,team2025kimi,sky_t1_2025,xu2025redstar,bespoke_stratos_2025}. While effective, these methods require substantial effort to collect and curate domain-specific data, raising the need for more effective and efficient training strategies. Thus, in the LLM reasoning space, there is a need for meta-learning: learning generalizable reasoning strategies that a model can rapidly adapt to multiple downstream tasks \cite{thrun1998learning,finn2017model,naik1992meta}. Following this motivation, we ask:

\textit{Can we train models that acquire general reasoning strategies and rapidly adapt them to new domains with minimal supervision?}\\

\begin{figure*}[t]
    \centering
    \includegraphics[width=0.7\linewidth]{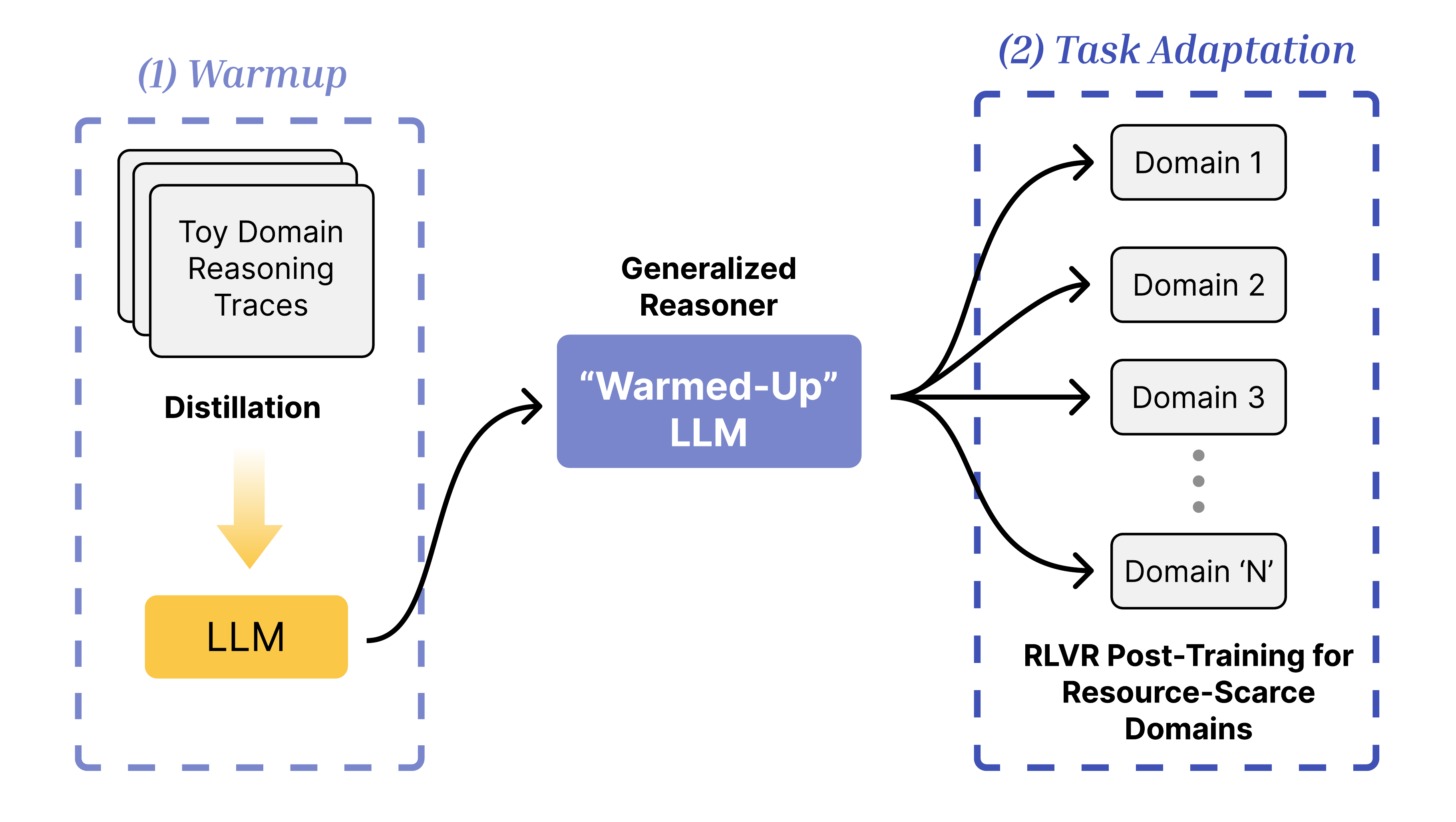}  
    \caption{Our two stage methodology involving a $(1)$ \textit{warmup} phase and a $(2)$ \textit{target-task adaptation} phase for training reasoning models in resource-constrained settings}
    \label{fig:methodology-fig}
\end{figure*}

In this work, we present a new framework for 
training reasoning models in resource-constrained settings, that is, when only a small amount of domain-specific training data is available. 
As shown in Figure \ref{fig:methodology-fig}, the framework has two training phases. 
In the first phase, we warm up the model on a simplified logic game environment—Knights \& Knaves \cite{smullyan1986name,xie2024memorization}—to induce generalizable reasoning behaviors. To accomplish this, we generate long CoTs for Knights \& Knaves questions, using QwQ-32B, a reasoning model \cite{qwen_qwq_32b_2025}. Then, without following any specialized data filtering scheme, we distill the responses into the base model to create a "warmed-up model". 
In the second phase, we adapt the warmed-up model to downstream tasks using only a small number of task-specific examples. This setup mimics the meta-learning paradigm: first learn abstract reasoning strategies that can generalize and quickly specialize to new tasks. The main contributions of the paper are:

\paragraph{Warmup alone creates a strong generalizable reasoning model.} In Section~\ref{warmup}, we show that the initial warmup phase alone yields a strong reasoning model. We warm up various models and consistently observe that it leads to significant improvement across benchmarks from different domains. For example, Qwen2.5-3B model with warmup gains $+10.2\%$ on MATH, $+15.3\%$ on HumanEval\textsuperscript{+}, and $+9.0\%$ on MMLU-Pro. Strikingly, warming up the Qwen2.5-14B model increases its MATH accuracy to $77.4\%$, closely approaching the $80.2\%$ achieved by full RLVR training with domain-specific data in a previous work \cite{zeng2025simplerl}. This suggests that the warmup phase activates generalizable reasoning behaviors in the model. Additionally, we find that our K\&K distillation often outperforms distillation with s1K dataset, a well-curated dataset of Long CoTs from diverse domains.

\paragraph{Warmup turns the model into an effective meta-learner}
In Section~\ref{adaptation-reasoning}, we show that the warmed-up model adapts more effectively when trained with RLVR using limited data. When both the base model and the warmed-up model are RLVR-trained on the same small dataset ($\leq100$ examples), the warmed-up model consistently outperforms the base model. For example, with the Qwen2.5-3B model, the warmed-up model achieves absolute gains of $+6.7\%$ on MATH and $+5.0\%$ on HumanEval\textsuperscript{+} compared to the base model RLVR-trained under the same conditions. Remarkably, on MATH, our two-stage approach on just 100 random examples matches the performance of a model trained directly with RLVR on the full training set of 7,500 examples. In addition to stronger final performance, the warmed-up model also converges faster, requiring fewer training steps to reach its peak. These findings demonstrate that the warmup phase turns the model into an effective meta-learner, enabling both efficient and effective RLVR training.

\paragraph{Warming up allows a model to maintain cross-domain generalizability even after subsequent RLVR training on a specific domain.}
RLVR training can sometimes lead models to adopt concise, domain-specific reasoning patterns that optimize for the target task \cite{fatemi2025concise}. While such specialization can yield strong in-domain performance, we show in Section \ref{cross-domain-generalization} that it often comes at a cost of degrading cross-domain reasoning abilities. However, 
by introducing the warmup phase into the pipeline, we can retain broader generalization capabilities even after task-specific RLVR training. To further assess generalization, we extend our evaluation beyond the commonly studied areas of math and coding to include subsets of the MMLU-Pro benchmark (in particular, physics \& history), offering a more comprehensive view of how RLVR affects performance in varied domains \cite{guo2025deepseek,team2025kimi,wei2025swe,hu2025open,zeng2025simplerl}. 

With regards to performance, sample efficiency, and generalization, our two-stage approach consistently outperforms baseline models, underscoring the effectiveness of simple warmup with domain-agnostic reasoning traces. These results highlight the promise of our method as a practical strategy for training robust reasoning models in resource-constrained settings.


\section{Warmup phase}
\label{warmup}
The warmup stage is based on the idea that reasoning behaviors—such as self-reflection and self-correction—are broadly applicable across domains. Therefore, this stage is designed to expose the model to such reasoning behaviors such that it can later apply them across various domains. While in past work such a warmup phase has mainly relied on domain-specific data \cite{luong2024reft,guo2025deepseek}, we instead train the LLM on reasoning traces from the simple Knights \& Knaves (K\&K) logic game \cite{smullyan1986name,xie2024memorization,xie2025logic}. In this section, we ask: \textit{Can warming up a base model with domain-agnostic reasoning patterns lead to substantial improvements on domain-specific tasks?}

\subsection{Knights \& Knaves logic game}

\begin{table}[h]
\centering

\begin{tabular}{p{0.9\linewidth}}
\toprule
\textbf{Knights \& Knaves Puzzle} \\
\midrule
\textbf{Question}\\
A very special island is inhabited only by knights and knaves. Knights always tell the truth, and knaves always lie. You meet four inhabitants: Luke, Liam, Matthew, and Ella.

Luke says: “Ella is a knave.”
Liam says: “Liam is a knight if and only if Luke is a knave.”
Matthew says: “Liam is a knave if and only if Ella is a knight.”
Ella says: “Matthew is not a knight.” \textit{Who is a knight and who is a knave?}\\

\addlinespace

\textbf{Answer} \\
Luke is a knave; Liam is a knight; Matthew is a knave; Ella is a knight
 \\
\bottomrule
\end{tabular}
\caption{An example Knights \& Knaves logic puzzle.}
\label{tab:knights-knaves-example}
\end{table}

In K\&K, the goal is to determine which characters are knights (who always tell the truth) and which are knaves (who always lie), based on their statements (Example shown in Table \ref{tab:knights-knaves-example}). We choose K\&K for two reasons. First, solving these problems requires extensive reasoning, which highlights generalizable reasoning strategies in the teacher model's responses. Second, K\&K relies solely on basic boolean logic and is independent of domain-specific knowledge—such as mathematical theorems or specialized software libraries—allowing the model to focus purely on reasoning behaviors rather than memorization of domain-specific content.


\subsection{Experiment Details}
\label{experiment-details}

We mainly conduct our experiments on the Qwen2.5-3B base model, chosen for its compact size and strong baseline abilities \cite{yang2024qwen2}.
 For warmup, we use QwQ-32B \cite{qwen_qwq_32b_2025}, a strong reasoning model, to generate long CoTs on K\&K questions. One example of such long CoT is shown in Table \ref{tab:knights-knaves-reasoning}. Since our goal is to collect responses reflecting extensive reasoning behaviors rather than improving accuracy on K\&K, we do not apply rejection sampling that filters out wrong responses, similar to the approach taken in some prior work \cite{gandhi2025cognitive, li2025llms}.



After generating the K\&K reasoning traces, we perform Supervised Fine-Tuning (SFT) on the base model with the traces to warmup our model. All details on SFT training with K\&K is available in Appendix \ref{distillation-appendix}. 

As evaluating LLMs can be noisy \cite{hochlehnert2025sober}, we sample multiple times for each question with careful answer extraction mechanisms, as discussed in Appendix \ref{evaluation-appendix}. 
We deal with three diverse datasets to assess the performance of our two-stage methodology across tasks with fundamentally different requirements:
\begin{enumerate}
    \item \textbf{MATH:}     A benchmark comprising 12,500 competition-level math problems (7,500 in train and 5,000 in test) designed to assess mathematical reasoning in LLMs \cite{hendrycks2021measuring}. For evaluation, we use the MATH500 subset of the test set \cite{lightman2023let}.
    \item \textbf{HumanEval\textsuperscript{+}:}     An extended benchmark of 164 hand-crafted programming problems, each with function signatures, docstrings, and unit tests, aimed at evaluating the code generation capabilities of LLMs \cite{chen2021evaluating,liu2023your}. In this section, we use all 164 questions for evaluation. When training with RLVR (Section 3), we randomly sample 50 questions and use the remaining examples for evaluation.
    \item \textbf{MMLU-Pro:}   An enhanced version of the Massive Multitask Language Understanding (MMLU) benchmark, featuring challenging questions across 14 categories like physics, law, history, and so on to test real-world text understanding and complex problem solving \cite{hendrycks2020measuring,wang2024mmlu}. For MMLU-Pro evaluation, we use the entire dataset except the examples we set aside for RLVR training (Section 3).\\
\end{enumerate}



\subsection{Results for Warmup Only}
\label{results-for-warmup}
\newcommand{\up}{\textcolor{green}{\ensuremath{\uparrow}}}
\newcommand{\down}{\textcolor{red}{\ensuremath{\downarrow}}}

\begin{table*}
  \centering
  \begin{tabular}{llll}
    \toprule
    \textbf{Model} & \textbf{MATH (\%)} & \textbf{HumanEval\textsuperscript{+} (\%)} & \textbf{MMLU-Pro (\%)} \\
    \midrule
    Qwen2.5-3B      & 43.8 ± 0.8            & 32.5 ± 1.2  & 29.2 ± 0.3  \\
    \quad K\&K  & \textbf{54.0 ± 1.4}~\up & \textbf{47.8 ± 3.7}~\up  & \textbf{38.2 ± 0.4}~\up   \\
    \quad s1K    & 53.9 ± 1.3~\up        & 39.5 ± 4.0~\up & 28.1 ± 0.1~\down   \\
    
    \addlinespace
    \midrule
    Qwen2.5-Math-1.5B      & 41.2 ± 1.6            & \textbf{8.8 ± 2.0}  & 18.3 ± 0.2  \\
    \quad K\&K  & \textbf{65.2 ± 1.0}~\up & 5.5 ± 0.8~\down  & \textbf{27.5 ± 0.2}~\up   \\
    \quad s1K    & 53.7 ± 1.1~\up        & 7.0 ± 0.7~\down & 22.4 ± 0.5~\up   \\

    \addlinespace
    DeepSeek-Math-7B-Base      & 20.6 ± 2.3            & 5.3 ± 1.2  & 21.1 ± 0.5  \\
    \quad K\&K  & 28.7 ± 2.1~\up & \textbf{13.8 ± 4.9} ~\up  & 23.2 ± 0.3 ~\up   \\
    \quad s1K    & \textbf{32.4 ± 2.5}~\up        & 5.9 ± 3.3   ~\up & \textbf{26.3 ± 0.6} ~\up   \\

    \addlinespace
    Qwen2.5-14B      & 55.6 ± 3.23            & 64.3 ± 3.0  & 52.7 ± 0.4  \\
    \quad K\&K  & \textbf{77.4 ± 1.9} ~\up  & \textbf{67.3 ± 2.3} ~\up  & \textbf{62.7 ± 0.2} ~\up   \\
    \quad s1K    & 51.5 ± 0.7 ~\down        & 24.3 ± 1.9   ~\down & 34.9 ± 0.7 ~\down   \\
    
    \bottomrule
  \end{tabular}
  \caption{Performance of the warmed-up model distilled with the K\&K game data and s1K data. Best performance highlighted in bold. We share results of warmup on other model families in Appendix~\ref{other-models}.}
  \label{performance-table}
\end{table*}

Table~\ref{performance-table} shows the performance of warmup for the three domains for four base models: the Qwen2.5-3B base model as discussed previously; the Qwen2.5-1.5B-Math, a math-specialized smaller model \cite{yang2024qwen2math}; DeepSeek-Math-7B-Base, a relatively weaker but larger model \cite{shao2024deepseekmath}, and Qwen2.5-14B, an even larger model from the Qwen family \cite{yang2024qwen2}. 
For comparison against our non-curated K\&K game dataset, we also distilled the base model with s1K, a dataset comprising of methodically collected, high-quality reasoning samples from various domains, including mathematics, physics, geography, logic, law, and so on \cite{muennighoff2025s1}. From Table~\ref{performance-table}, we observe:

\begin{enumerate}
    \item For the the Qwen2.5-3B base model, 
    although the warmup process does not convey any mathematics or domain-specific knowledge, \emph{warmup leads to marked improvements over the base model across the three domain-specific benchmarks}. This suggests that the warmup phase, using reasoning traces from the K\&K domain alone, can effectively activate general reasoning capabilities in the model. 
    \item In similar vain to the 3B model, we see strong improvements for the other base models as well. In case of the 1.5B model, we see significant improvements on the MATH dataset, possibly because of the math-specialized nature of this model. Similarly, for the 14B Qwen model, we find significant improvements across all three benchmarks. Notably, the model achieves $77.4\%$ on MATH after warmup alone, a result that approaches the $80.2\%$ reported in prior work using full-scale RL training with domain-specific data \cite{zeng2025simplerl}. In case of the DeepSeek family model as well, we see notable improvements across the benchmarks, showcasing the effectiveness of domain-agnostic warmup.
    \item As shown in the "s1K" rows of Table~\ref{performance-table} for the models, \emph{distilling  with CoTs for simple logic games (i.e., warming up) generally matches or even performs better than distilling with domain-knowledge-rich s1K data.} Also, Qwen2.5-14B experiences notable degradation after s1K training. Thus, we explore a different set of hyperparameters for optimal training but still find minimal gains (see Appendix~\ref{overfitting}).
\end{enumerate}




Some recent studies have showcased that distillation with reasoning-intensive Long-CoT (like s1K data) on smaller models might struggle to learn effectively, showcasing minimal gains or even degradation \cite{xu2025phi,li2025small}. Our findings show that distillation in a simplified, domain-agnostic setting with non-curated data like K\&K may actually benefit smaller LLMs by avoiding the complexity of domain-specific reasoning and leading to general performance gains. We confirm the validity of the warmup methodology by distilling on models from other model families as well (see Appendix~\ref{other-models}). Furthermore, to assess robustness to traces from a different teacher model, we do an ablation study by using DeepSeek-R1 traces for distillation and continue to observe strong performance gains (see Appendix~\ref{r1-distillation}).

\subsection{Understanding the Effect of Reasoning Behaviors}


The reasoning traces generated by the reasoning teacher model (QwQ) using the K\&K games data have general reasoning behaviors, like self-reflection and verification, along with task-specific logic like performing boolean operations and handling states. To test whether the performance gains stem from the reasoning behaviors rather than from the domain-specific logic of Knights \& Knaves (K\&K), we run a controlled experiment. 

In addition to the warmup experiment just described in Section \ref{results-for-warmup}, we perform another distillation experiment. We again choose the Qwen2.5-3B model as our base model but now perform distillation with short CoTs which don't have explicit reasoning behaviors. To collect the short CoTs, we use the non-reasoning Qwen2.5-32B as the teacher model. In this setting, the model quickly overfits to the K\&K task, exhibiting significantly reduced generalization abilities. For instance, accuracy on MATH drops to just $11\%$. In contrast, our "warmed-up" model improves to $54\%$ as shown in Table \ref{performance-table}. This shows that the model trained on the K\&K games \emph{with} reasoning supervision successfully abstracts transferable reasoning patterns, leading to substantial performance gains across diverse tasks.

\section{Target-Domain Adaptation}
\label{adaptation}
While simple reasoning behaviors may generalize across domains, as demonstrated in the previous section, many downstream tasks require domain-specific reasoning, e.g., decomposing complex numerical problems in math or generating unit tests in coding \cite{sky_t1_2025}. Such skills need to be polished and are unlikely to emerge from training on a toy domain such as Knights \& Knaves alone. Therefore, we introduce a target adaptation phase, where the warmed-up model is RLVR-trained using a few samples ($\leq100$) from the resource-constrained target domain.

We first ask \emph{can an initial warmup phase with domain-agnostic reasoning traces give a significant boost to the performance of RLVR training?} If the answer is yes, then in the spirit of meta-learning, we can simply warm up the base model once and then adapt it to many different low-resource domains with RLVR. 

\begin{figure*}[t]
    \centering
    \includegraphics[width=0.9\linewidth]{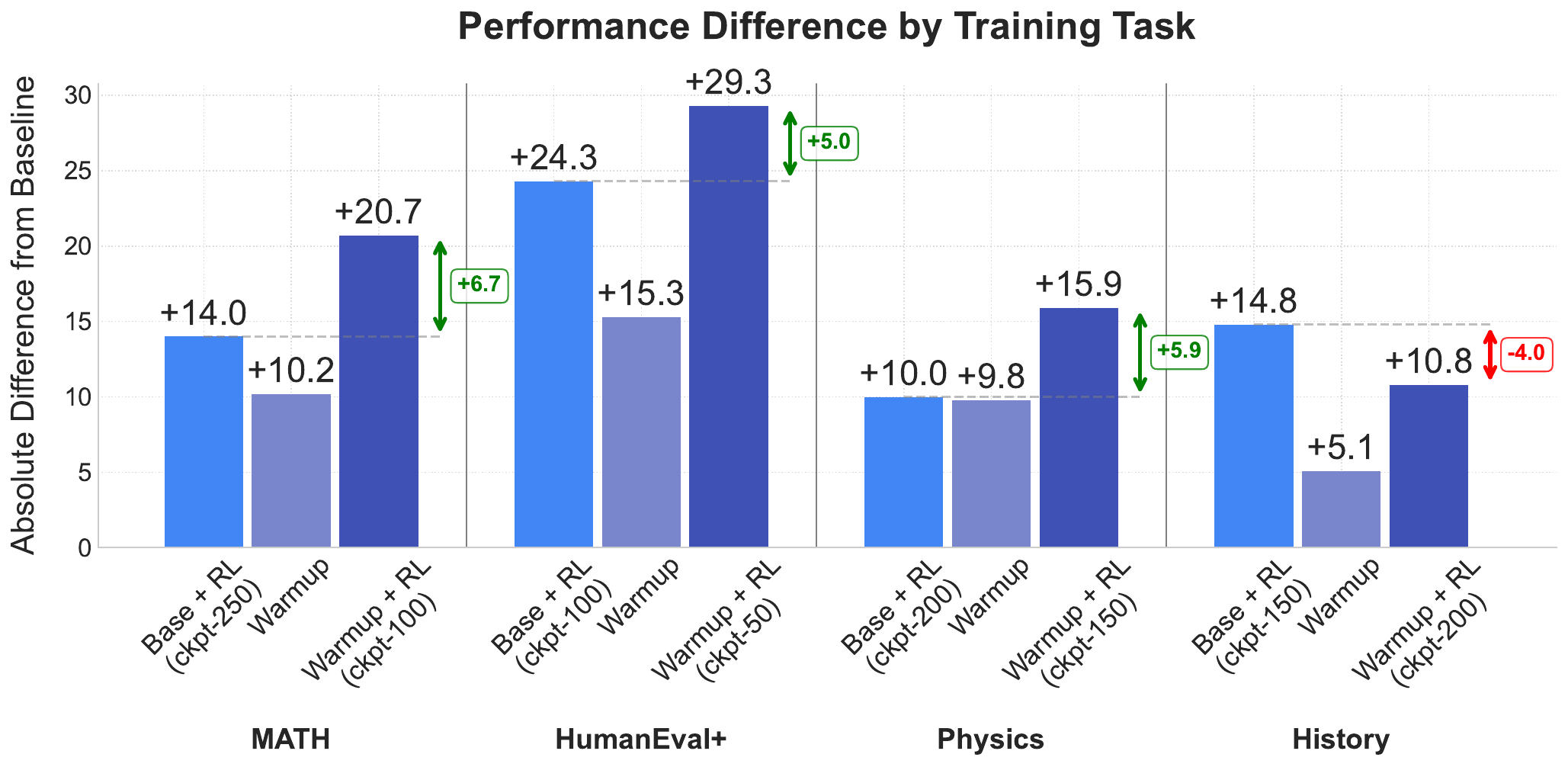}  
    \caption{Absolute percentage increase relative to the base model for MATH, HumanEval\textsuperscript{+}, and MMLU-Pro subsets (Physics \& History) for various models: RLVR on base model, warmup only, and RLVR on warmed-up model}
    \label{fig:adaptation-results}
\end{figure*}

\subsection{Adaptation for math and coding}
\label{adaptation-reasoning}
We first conduct experiments across the MATH and HumanEval\textsuperscript{+} datasets discussed in Section \ref{experiment-details}.
To create the resource-constrained dataset, we randomly sample 100 questions from the MATH train set and 50 samples (of 164 available) from the HumanEval\textsuperscript{+}. We use the remainder of the dataset for evaluation in case of HumanEval\textsuperscript{+}, and we use the MATH500 dataset for evaluation \cite{lightman2023let} for MATH. We perform two RLVR runs per dataset, one on the base model and one on our warmed-up (that is, K\&K distilled) model.
For our experiments, we employ the unbiased GRPO algorithm (Dr. GRPO)\cite{liu2025understanding}. Due to the high cost of RL training, we perform experiments only on the Qwen2.5-3B base model.


\paragraph{MATH}
Using the training set of only 100 examples, we perform RLVR fine-tuning on both the base model and the warmed-up model, keeping all hyperparameters constant. As shown in first column of Figure~\ref{fig:adaptation-results}, the base model after RLVR training achieves an absolute improvement of $+14.0\%$ over the base model after 250 training steps. In comparison, the warmed-up model
achieves a significantly larger absolute gain of $+20.7\%$  after just 100 steps. 


Since we have a large training set (7,500 examples) for MATH, in order to assess performance at scale, we also conduct RLVR training on the base model with the full MATH training set until the model's performance plateaus. Interestingly, the final performance of the base model trained on the full dataset is comparable to that of the warmed-up model trained on just 100 examples: the former reaches a score of $63.2\%$ on MATH500, while the latter achieves $64.5\%$. While large-scale training with well-curated datasets remains valuable and may ultimately lead to higher ceilings, these results highlight that a warmed-up model can achieve strong performance with fewer examples—demonstrating greater performance and sample efficiency in low-resource settings. Detailed results are shown in Table \ref{math-results} and \ref{comp-math-results} in the Appendix.


\paragraph{HumanEval+} For this experiment, we use the RLVR training set of 50 examples. As shown in Figure~\ref{fig:adaptation-results}, the base model achieves an absolute improvement of $+24.3\%$ after $100$ RLVR training steps. In comparison, the warmed-up model alone has good performance, showing an absolute gain of $+15.3\%$  without any domain-specific fine-tuning, and reaching an absolute improvement of $+29.3\%$ after just $50$ RLVR training steps. These results highlight the effectiveness of our two-stage approach in the coding domain as well. Detailed results are shown in Table \ref{coding-results} and \ref{comp-coding-results} in the Appendix.

\subsection{Adaptation for MMLU-Pro}

MATH and HumanEval\textsuperscript{+} primarily consist of problems that demand intensive reasoning \cite{hendrycks2021measuring,liu2023your}. To assess broader applicability, we expand our study to the MMLU-Pro benchmark. MMLU-Pro presents a broader mix, combining both challenging reasoning tasks and factual recall questions. For instance, MMLU-Pro includes questions from the physics domain, which require a mix of analytical problem-solving skills and familiarity with certain domain-specific concepts. It also includes the history domain, which has questions that rely more heavily on factual world knowledge. We perform RLVR runs on these two specific subsets (physics and history) to assess how the model performs in diverse reasoning and knowledge demands. 

\paragraph{Physics} From the 1,299 available examples in the physics subset, we randomly sample 100 instances for training and use the remaining for evaluation. As shown in Figure~\ref{fig:adaptation-results}, RLVR training improves the base model’s absolute performance by $+10.0\%$, reaching peak performance after 200 steps. In comparison, the warmed-up model already exhibits strong performance out of the box with an absolute $+9.8\%$  gain, and further improves to $+15.9\%$ after just 150 training steps, demonstrating both faster convergence and higher final performance. We find in this case where we have a mix of both reasoning-intensive and knowledge-intensive questions, RLVR training still leads to significant gains. Detailed results are shown in Table \ref{physics-results} and \ref{comp-physics-results} in the Appendix.

\begin{figure*}[t]
    \centering
    \includegraphics[width=0.9\linewidth]{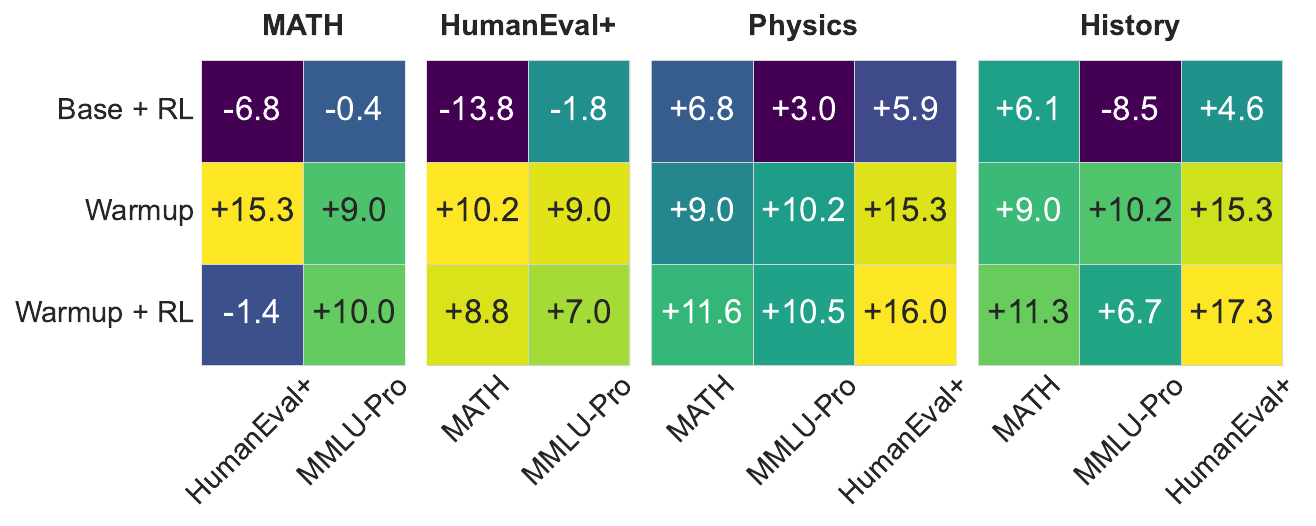}  
    \caption{Generalization results for base+RLVR, warmup-only, and warmup+RLVR models. Top labels indicate RLVR training domain; bottom labels indicate evaluation domain. Warmup uses no domain-specific data.}
    \label{fig:generalization-results}
\end{figure*}

\paragraph{History} The history subset of MMLU-Pro contains 381 examples, from which we select 100 at random for training and reserve the rest for evaluation. For this domain, RLVR training applied directly to the base model achieves a notable absolute improvement of $+14.8\%$, outperforming the warmed-up model, which reaches a lower gain of $+10.8\%$ after RLVR training. Past work has shown that for knowledge-intensive tasks, even direct prompting without step by step reasoning outperforms Chain of Thought prompting \cite{wei2022chain,wang2024mmlu}. Thus, for knowledge-intensive tasks, having a short CoT might be the optimal policy. As shown in Table \ref{tab:history-example-long}, the base model after RLVR training learns to produce very short responses on history questions, hence giving better performance. In comparison, since the warmed-up model has been primed for reasoning through the warmup phase, it struggles to perform on par with the base model on a non-reasoning task like history.
However, as we discuss in Section~\ref{cross-domain-generalization}, this domain-specific improvement with RLVR directly applied to the base model may come at the cost of broader generalization. Detailed results for the history runs are shown in Table \ref{history-results} and \ref{comp-history-results} in the Appendix.
\\
\newline
While previous work has found that SFT using short CoT can limit the benefits of subsequent RL ~\cite{zeng2025simplerl}, in this section, we showed that RL training a model warmed up with long CoT (reasoning traces), although from a simplified domain (K\&K), can significantly improve generalization and sample efficiency. This is especially beneficial for training in resource-constrained settings with limited samples. Details on all hyperparameters related to our experiments, training curves, and results along with standard deviations are available in Appendix \ref{rlvr-experiment-appendix}.


\begin{figure}[h]
  \includegraphics[width=\columnwidth]{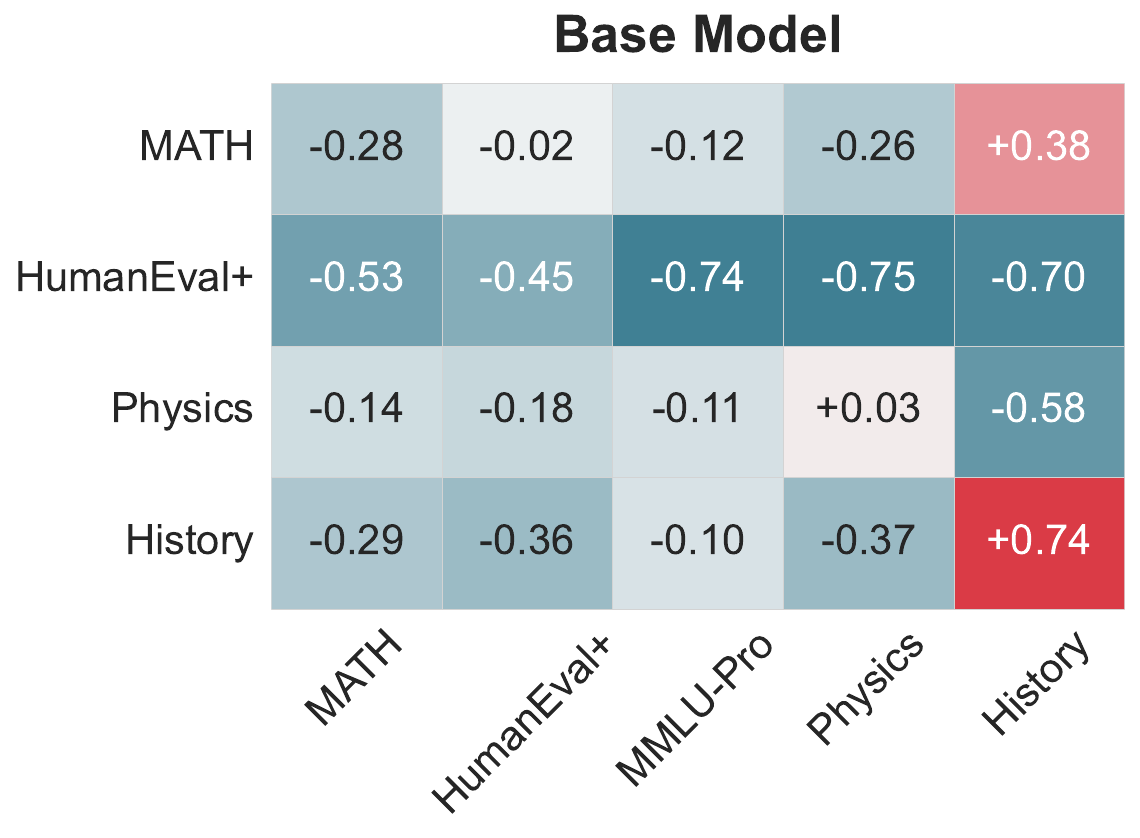}
  \includegraphics[width=\columnwidth]{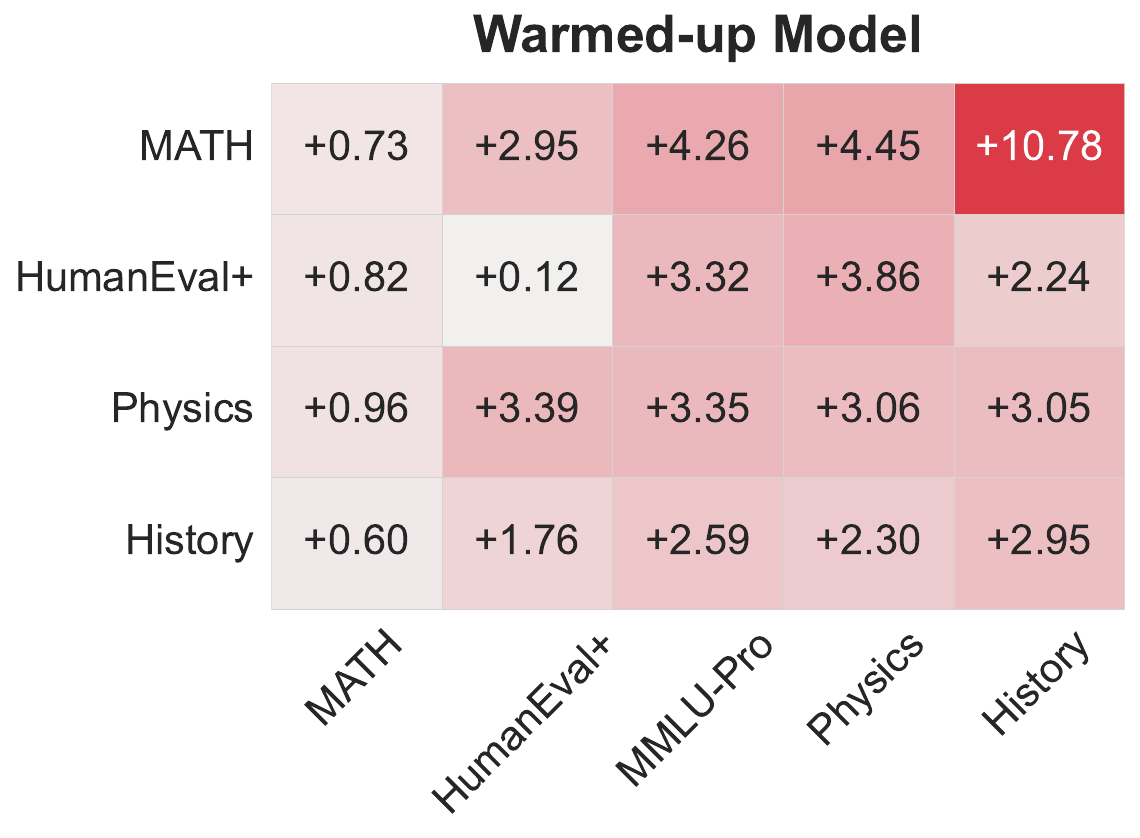}
  \caption {Relative change in completion length (vs. base model) after training on each dataset (y-axis) and evaluating on others (x-axis). Top: base model; Bottom: warmed-up model.}
\label{fig:completions-heatmap}
\end{figure}

\subsection{Cross-Domain Generalization}
\label{cross-domain-generalization}

While recent research has shown that large scale RLVR-trained models can generalize to out-of-domain tasks \cite{zeng2025simplerl,hu2025open,xie2024memorization}, we study generalization across multiple distinctly different domains when the sample size is low.
We test for generalization across domains with fundamentally different structures, such as math, code generation, and language understanding as discussed in Section \ref{experiment-details}. 

In this section, we compare generalizability across the two RLVR runs discussed in Section \ref{adaptation}; one with the base model and the other with the warmed-up model. For a model RLVR-trained on one domain (such as physics), we test generalization by evaluating on other domains (such as mathematics and coding). We ask the question,
\emph{how does the generalization performance of the two runs compare? Does warming up the base model before RLVR training help to maintain or even improve generalization to other domains?}

\paragraph{Loss in Generalizability on Base Model} As shown in Figure \ref{fig:generalization-results}, we observe that although RLVR with domain-specific questions typically increases the performance of the model when asked to solve problems in the same domain, it tends to lose its initial base-model reasoning capability when asked to solve problems in different domains. This effect is especially pronounced in case of training on reasoning-intensive domains like MATH and HumanEval\textsuperscript{+}.
For instance, RLVR training the base model on HumanEval\textsuperscript{+} leads to a $-13.8\%$ drop on MATH, while RLVR training on the History subset results in a $-8.5\%$ drop on MMLU-Pro. This degradation suggests that RLVR training may push the model to internalize narrow, domain-specific heuristics, thereby weakening its broader reasoning abilities. This interpretation is further supported by our observation (Appendix \ref{rlvr-experiment-appendix}) that RLVR fine-tuning often leads to more concise completions—consistent with prior findings \cite{fatemi2025concise, kim2025reinforcement}—potentially reflecting a contraction in reasoning depth.

\paragraph{Improved Generalizability with Warmup} 
Figure \ref{fig:generalization-results} also shows the generalization results when we begin the pipeline with the warmed-up model. 
While the forgetting effect is occasionally observed in the warmed-up model (e.g., a $-1.4\%$ decrease on HumanEval\textsuperscript{+} relative to the base model after RL training on MATH), \emph{beginning with the warmed-up model rather than with the base model consistently yields stronger cross-domain generalization.}

Figure \ref{fig:completions-heatmap} shows the change in average completion length relative to the base model after RLVR training on different datasets and evaluating on all others. Although both models have short completion lengths after training(\ref{rlvr-experiment-appendix}), the warmed-up model has a strictly higher completion length relative to the base model's original completions.
The consistently longer completions may help preserve the model's general reasoning abilities across multiple domains. 

Interestingly, we find that training on a domain like Physics, which blends both multi-step reasoning and factual recall, better maintains cross-domain performance for both the base and warmed-up models. This observation suggests that task diversity during RLVR training may play a critical role in preserving generalization.





\section{Discussion \& Related Work}
\label{gen_inst}

\subsection{Meta-Learning \& Generalization}

Previous studies have adopted two main techniques for developing reasoning LLMs: distillation with well-curated long CoTs, and large scale RLVR training. Both techniques demand large amounts of high-quality data \cite{guo2025deepseek,hu2025open,muennighoff2025s1}. These methods typically focus on task-specific optimization and do not directly address how to equip models with the ability to adapt rapidly to new domains under low-resource scenarios. Our work, inspired by the meta-learning paradigm, aims to fill this gap \cite{thrun1998learning,naik1992meta,finn2017model}.

While some studies have shown impressive improvements in math domains with RLVR training on tiny curated datasets  \cite{wang2025reinforcement,fatemi2025concise}, our approach differs in both scope and methodology. We prioritize sample efficiency and cross-domain generalization across diverse domains, with fundamentally different requirements (math, coding, and language understanding), without relying on selective filtering during either the warmup or downstream adaptation phases.



\subsection{SFT Warmup for Reasoning}
Previous work has shown that reasoning behaviors such as self-verification, backtracking, and self-reflection play a critical role in effective reasoning for large language models (LLMs) \cite{gandhi2025cognitive,openai_o1,guo2025deepseek}. Although a base model might exhibit some reasoning behaviors already, performing distillation with long CoTs from tasks requiring extensive reasoning, like Olympiad level math problems or complex coding tasks, can enable these models to become effective reasoners \cite{muennighoff2025s1,zhao2025echo,liu2025understanding,sky_t1_2025,bespoke_stratos_2025}. However, rather than relying on meticulously curated domain-specific data, we explore the effectiveness of distilling reasoning patterns in the absence of such meticulously curated, domain-specific Long CoTs. We find that warmup with simple domain-agnostic reasoning may also serve as a strong activation mechanism that helps the model to learn to piece together reasoning behaviors for application in multiple downstream domains. 

Some previous studies have focused on learning generalizable behaviors in the form of formal logic and evaluating their transferability across various domains \cite{morishita2023learning,morishita2024enhancing}. Our motivation is similar, but we instead analyze the transfer of general reasoning skills, such as self-verification and self-reflection, that have been adopted in recent reasoning LLMs \cite{guo2025deepseek,openai_o1}. Moreover, we not only study the generalizability of these reasoning skills but also investigate how they improve sample efficiency and downstream performance after RlVR training.

\section{Conclusion}

We present a two-stage training framework that combines a lightweight warmup phase with reinforcement learning using verifiable rewards (RLVR) to develop reasoning-capable LLMs under limited supervision. By distilling general reasoning from a toy domain and fine-tuning on a small set of target examples, our approach consistently outperforms direct RLVR on base models across math, code, and language understanding tasks. Warmup enhances sample efficiency, boosts performance, and preserves cross-domain generalization—highlighting its promise as a practical strategy for training robust LLMs in low-resource settings.

\section*{Limitations}
\label{limitations}
Our findings demonstrate that reasoning distilled from a simple, abstract domain like Knights \& Knaves (K\&K) can generalize effectively to tasks in math, code generation, and general language understanding—underscoring the promise of domain-agnostic reasoning as a transferable prior. However, it remains an open question how well such reasoning transfers to more complex tasks involving richer dynamics, interaction, or domain-specific constraints, like multi-agent environments.

While we selected K\&K for its simplicity and interpretability, we do not claim that it is the only source of transferable reasoning. Other synthetic environments may offer equally or more effective reasoning priors tailored to downstream needs.

We encourage future work to explore the design of such environments as playgrounds for models to learn generalizable reasoning behaviors and apply them to specific domains. Additionally, limited by our compute, we were able to perform distillation only till 14B models and RLVR train on a 3B model; evaluating this methodology on larger models could offer further insights into the scalability and generalization potential of warmup-based reasoning.

\section*{Acknowledgements}
This work is submitted in part by the NYU Abu Dhabi Center for Artificial Intelligence and
Robotics, funded by Tamkeen under the Research Institute Award CG010. Some experiments were
carried out on the High Performance Computing resources at New York University Abu Dhabi.

\bibliography{acl_latex}

\appendix



\section{Appendix}
\label{appendix}

\subsection{The Knights \& Knaves (K\&K) Game}
\label{distillation-appendix}
We use Knights \& Knaves questions from a previous work \cite{xie2024memorization}. The questions deal with 3 to 7 characters. Here, the number of characters determines the complexity of the problem. The dataset has a total of $5,000$ examples. We retain $90\%$ of the dataset $(4,500)$ for training and keep the rest for evaluation $(500)$. After distillation on the Qwen 3B model, performance on K\&K improves from $19.4\% \pm 1.6$ to $81\% \pm 1.5$ on the test set.

In Table \ref{tab:knights-knaves-example}, we show an example of the K\&K game. In Table \ref{tab:knights-knaves-reasoning}, we show an example of a reasoning trace, generated by the reasoning model QwQ-32B, on one of such puzzles.

\onecolumn

\begin{longtable}{p{0.9\textwidth}}
\toprule
\textbf{Prompt} \\
\midrule

A conversation between User and Assistant. The user asks a question, and the Assistant solves it.
The assistant first thinks about the reasoning process in the mind and then provides the user with the answer.
The reasoning process and answer are enclosed within <think> </think> and <answer> </answer> tags, respectively.
i.e., <think> reasoning process here </think> <answer> answer here </answer>.
User: Please solve this logical reasoning problem. After thinking, when you finally reach a conclusion, clearly state the identity of each character within <answer> </answer> tags i.e., <answer> (1) Zoey is a knight
(2) ... </answer>. Here is my question,

\textit{A very special island is inhabited only by knights and knaves. Knights always tell the truth, and knaves always lie. You meet 3 inhabitants: Riley, Emily, and Lucas. In Riley's words: "Lucas is not a knave". Emily said that Lucas is a knight and Riley is a knight. In a statement by Lucas: "Emily is a knight or Emily is a knave". So who is a knight and who is a knave?}

Assistant: <think>\\
\addlinespace
\midrule

\textbf{Reasoning from QwQ-32B} \\
Okay, so I need to figure out who among Riley, Emily, and Lucas are knights or knaves. Let me start by recalling that knights always tell the truth and knaves always lie.
\\
First, let's note down each person's statements:

- Riley says: "Lucas is not a knave."
- Emily says: "Lucas is a knight and Riley is a knight."
- Lucas says: "Emily is a knight or Emily is a knave."

\\
Hmm, Lucas's statement is a bit tricky. Let's break it down. The statement "Emily is a knight or Emily is a knave" is actually a tautology because every inhabitant is either a knight or a knave. So regardless of whether Emily is a knight or knave, that statement is always true. Wait, but if Lucas is a knave, he must lie. But his statement is a true statement, so if he's a knave, he can't tell the truth. That means Lucas cannot be a knave, because his statement is always true. Therefore, Lucas must be a knight? Because if he were a knave, he would have to lie, but his statement is inherently true, so he can't lie about it. So Lucas must be a knight.
\\
Wait, let me think again. Let's analyze Lucas's statement more carefully. The statement "Emily is a knight or Emily is a knave" is a logical OR. Since every inhabitant is either a knight or a knave, this OR is always true. So the statement is a tautology. Therefore, regardless of Emily's actual status, the statement is true. So if Lucas is a knight, he tells the truth, so that's fine. If he were a knave, he would have to lie. But since his statement is true, a knave cannot say it. Therefore, Lucas must be a knight. That's solid.
\\
So Lucas is a knight. Therefore, Riley's statement that "Lucas is not a knave" is true. Since Riley's statement is true, Riley must be a knight as well. Wait, but let's check Emily's statement. Emily says that Lucas is a knight and Riley is a knight. Since we just concluded both are knights, then Emily's statement is true. Therefore Emily must also be a knight. But that can't be right because then all three would be knights, but let me check again.
\\
Wait, hold on. Let me go step by step.
\\
First conclusion: Lucas is a knight (from his own statement being a tautology). Therefore, Riley's statement "Lucas is not a knave" is true, so Riley is a knight. Then Emily's statement is "Lucas is a knight and Riley is a knight." Since both parts are true (Lucas and Riley are knights), Emily's statement is true, so Emily must be a knight. That would mean all three are knights. But is there any contradiction here?
\\
Wait, but let me see. If all three are knights, then all their statements are true, which they are. So that seems okay. But maybe I missed something? Let me try another approach.
\\
Alternative approach: Suppose Lucas is a knave. Then his statement "Emily is a knight or a knave" would be a lie. But since the statement is a tautology (always true), lying about it would require the statement to be false. But since it's always true, Lucas can't be a knave. Therefore, Lucas must be a knight. So that part holds.
\\
Therefore, Lucas is a knight. Then Riley's statement is true, so Riley is a knight. Then Emily's statement is "Lucas is a knight and Riley is a knight." Since both are true, Emily's statement is true, so Emily is a knight. So all three are knights. That seems possible, but maybe the problem expects some of them to be knaves. Let me check again for errors.
\\
Wait, perhaps I made an error in analyzing Emily's statement. Let me parse Emily's statement again. She says "Lucas is a knight and Riley is a knight." If both are true, then Emily is a knight. If either is false, then her statement is false, so she would be a knave.
\\
But according to our previous conclusion, both are true, so she must be a knight. So that's consistent. Therefore all three are knights. But the problem says "inhabited only by knights and knaves," so having all knights is allowed. Hmm, but maybe I missed something else?
\\
Alternatively, perhaps the problem has a trick. Let me consider if there's another possibility. Suppose Lucas is a knight (as before). Then Riley is a knight. Emily's statement is true, so she's a knight. So all three are knights. That works. 
\\
Alternatively, is there a scenario where someone is a knave?
\\
Suppose, for contradiction, that Emily is a knave. Then her statement "Lucas is a knight and Riley is a knight" would be a lie. The negation of "A and B" is "not A or not B". So either Lucas is a knave, or Riley is a knave, or both. But we already established that Lucas must be a knight (since his own statement is a tautology). Therefore, if Emily is a knave, then Riley must be a knave. But Riley's statement "Lucas is not a knave" would then be a lie, meaning Lucas is a knave. But that contradicts our earlier conclusion that Lucas must be a knight. Therefore, Emily can't be a knave. So Emily must be a knight.
\\
Therefore, all three are knights. Wait, but that seems a bit too straightforward. Let me check again.
\\
Alternatively, perhaps the problem is designed so that one is a knave. Let me see:
\\
Wait, Lucas's statement: "Emily is a knight or a knave" — since every inhabitant is one or the other, that's always true, so Lucas must be a knight. So that part is solid. Then Riley's statement is true, so Riley is a knight. Emily's statement is true, so she's a knight. So all three are knights. So the answer would be all three are knights. 
\\
Hmm, but maybe the problem is expecting a different answer? Let me see the original problem again:
\\
Problem states: Riley says Lucas is not a knave. Emily says "Lucas is a knight and Riley is a knight." Lucas says "Emily is a knight or Emily is a knave."
\\
Wait, the problem might have a trick in Emily's statement. Let me parse Emily's exact words again: "Lucas is a knight and Riley is a knight." So if Emily is a knave, then her statement must be false, so either Lucas is not a knight (so knave) or Riley is not a knight (so knave). But since we know Lucas is a knight, that would require Riley to be a knave. But Riley's statement is "Lucas is not a knave," which would be true (since Lucas is a knight), so Riley would be a knight, contradicting the assumption that Riley is a knave. Therefore Emily can't be a knave. 
\\
Therefore, all three must be knights. 
\\
Alternatively, is there any other possibility? Let me try assuming that Emily is a knave again. If Emily is a knave, then her statement is false. Since the statement is "Lucas is a knight AND Riley is a knight," the negation is "Lucas is a knave OR Riley is a knave." But since Lucas must be a knight (from his own statement), then the only way for Emily's statement to be false is if Riley is a knave. But then Riley's statement "Lucas is not a knave" would be true, so Riley would be a knight, which contradicts the assumption that Riley is a knave. Therefore Emily can't be a knave. 
\\
Therefore the only possible solution is that all three are knights. 
\\
Wait, but the problem says "a very special island inhabited only by knights and knaves." There's no rule that there must be at least one of each, so all three being knights is acceptable. 
\\
Hmm, so the conclusion is that all three are knights. But maybe I made a mistake in the analysis. Let me try another angle.
\\
Suppose Lucas is a knight (as established). Therefore, his statement is true, so the rest follows. Then Riley's statement is true, so Riley is a knight. Emily's statement is true, so Emily is a knight. So yes, that's consistent. 
\\
Alternatively, let's suppose that Lucas is a knave. Then his statement "Emily is a knight or a knave" is a lie. But since the statement is a tautology (always true), lying about it is impossible. So Lucas cannot be a knave. Therefore, Lucas must be a knight. 
\\
Thus, the only possible conclusion is all three are knights. 
\\
But maybe the problem is designed so that someone is a knave? Let me think again.
\\
Wait, perhaps there's a mistake in considering that "Emily is a knight or a knave" is a tautology. Wait, actually, in logic, "A or not A" is a tautology, but here Emily is either a knight or a knave, so "Emily is a knight or Emily is a knave" is indeed a tautology. Therefore, Lucas's statement is always true, so he must be a knight. 
\\
Therefore, I think the correct answer is that all three are knights. So Riley: knight, Emily: knight, Lucas: knight. 
\\
Wait, but let me check once more. Let me tabulate possible scenarios:
\\
Case 1: All knights. Then all statements are true. That works.
\\
Case 2: Suppose Lucas is a knight, Riley a knight, Emily a knave. Then Emily's statement would be false, but since both parts are true (Lucas and Riley are knights), her statement is true, so she can't be a knave. So this case invalid.
\\
Case 3: Suppose Emily is a knave, then her statement is false, so either Lucas is knave or Riley is knave. But Lucas must be a knight, so Riley must be knave. But then Riley's statement "Lucas is not a knave" is true, making Riley a knight, contradiction. So Emily can't be a knave.
\\
Case 4: Suppose Riley is a knave. Then his statement "Lucas is not a knave" is false, so Lucas is a knave. But Lucas's statement would then be a lie, but his statement is a tautology (true), so Lucas can't be a knave. Contradiction. So Riley must be a knight.
\\
Thus, all three must be knights. Therefore the answer is all three are knights.
</think>
\\
<answer>
(1) Riley is a knight  
(2) Emily is a knight  
(3) Lucas is a knight  
</answer>
 \\
\bottomrule
\caption{Example of a reasoning trace for the Knights \& Knaves logic puzzle generated with QwQ-32B teacher model. As seen here, the response is quite verbose with complex reasoning behaviors like self-reflection, hypothesis testing, and so on.} 
\label{tab:knights-knaves-reasoning} \\
\end{longtable}

\twocolumn

\subsection{Codebase}
\label{codebase}
We use the TRL package\footnote{\url{https://huggingface.co/docs/trl/en/index}} with the DeepSpeed ZeRO Stage 3 \cite{rajbhandari2020zero} configuration for Supervised Fine-tuning and Reinforcement Learning with GRPO \cite{shao2024deepseekmath}.

The complete codebase for this work is available at \url{https://anonymous.4open.science/r/warmup-before-you-train-0EEF/}.

\subsection{Evaluation}
\label{evaluation-appendix}
We keep the temperature 0.7 and top-p 0.95 for both training and evaluation. Given the sensitivity during evaluation, we sample 4 times for each question and report the average accuracy along with the standard deviation.

We use the \texttt{math\_verify} package for math evaluation. For HumanEval\textsuperscript{+}, we base our evaluation on OpenAI's codebase\footnote{\url{https://github.com/openai/human-eval}}. Finally, for MMLU-Pro, we adapt our evaluation from MMLU-Pro's official repository\footnote{\url{https://github.com/TIGER-AI-Lab/MMLU-Pro}}. We use the \texttt{vllm} package for running inference for evaluation on H100 GPUs.\footnote{\url{https://docs.vllm.ai/en/latest/}}
All evaluation code is made available in the shared repository.

\subsection{Distillation Experiment Details}
\label{sft-hyperparameters}
For all distillation experiments, we used the hyperparameters as shown in Table \ref{tab:sft-hyperparams}.

\begin{table}[h]
  \centering
  \begin{tabular}{ll}
    \toprule
    \textbf{Hyperparameter}       & \textbf{Value} \\
    \midrule
    Optimizer                     &  AdamW \\
    Weight Decay                 &  $1e-4$ \\
    Warmup Steps         &  25 \\
    Max Sequence Length          &  32,768 \\
    Batch Size (per device)      &  1 \\
    Gradient Accumulation Steps  &  2 \\
    Number of Epochs             &  3 \\
    Learning Rate Scheduler      &  constant \\
    bf16              &  True \\
    GPUs              &  $6$ H100s \\
    \bottomrule
  \end{tabular}
\caption{Key hyperparameters used in supervised fine-tuning}
  \label{tab:sft-hyperparams}
\end{table}

For learning rate, we choose $1e-6$ for distillation with K\&K, and choose $2e-5$ for distillation with s1K data as they showed the best results in empirical observations.




\subsubsection{Warmup with K\&K on Other Models }
\label{other-models}
Recent studies have shown that Qwen model families show "unsual" behaviors, bringing into question whether patterns observed in Qwen models are generally applicable \cite{shao2025spurious}. Thus, to further confirm the validity of our approach, we experiment with warmup on other model families as well as shown in Table~\ref{other-performance-table}. We find that our warmup with K\&K data brings strong improvement in non-Qwen model families as well. 

\begin{table*}
  \centering
  \begin{tabular}{llll}
    \toprule
    \textbf{Model} & \textbf{MATH (\%)} & \textbf{HumanEval\textsuperscript{+} (\%)} & \textbf{MMLU-Pro (\%)} \\
    \midrule
    
    Olmo2-1B      & 4.6 ± 0.4  & 0.0  & 11.7 ± 0 \\
    \quad K\&K  & \textbf{5.5 ± 0.5}~\up & 0.0 & \textbf{11.0 ± 0.3} ~\down \\

    \addlinespace
    Olmo2-7B      & 14.3 ± 0.7 & 3.9 ± 0.5  & 20.0 ± 0.5 \\
    \quad K\&K  & \textbf{21.4 ± 1.1} ~\up & \textbf{4.8 ± 1.1} ~\up & \textbf{25.1 ± 0.1} ~\up \\

    \addlinespace
    Llama-3.2-3B      & 6.8 ± 0.3  & 0.9 ± 0  & 16.2 ± 0.6 \\
    \quad K\&K  & \textbf{8.5 ± 1.5} ~\up & \textbf{7.5 ± 1.5} ~\up & \textbf{18.3 ± 0.3} ~\up \\

    \addlinespace
    Llama-3.1-8B      & 11.8 ± 0.3  & 13.1 ± 0.6  & 23 ± 0.2 \\
    \quad K\&K  & \textbf{25.2 ± 0} ~\up & \textbf{17.1 ± 1.9} ~\up & \textbf{33.9 ± 0.7} ~\up \\
    \bottomrule
  \end{tabular}
  \caption{Performance of non-Qwen model families distilled with the K\&K game data. Best performance highlighted in bold.}
  \label{other-performance-table}
\end{table*}

\subsubsection{Overfitting}
\label{overfitting}
\begin{figure}[h]
  \includegraphics[width=\columnwidth]{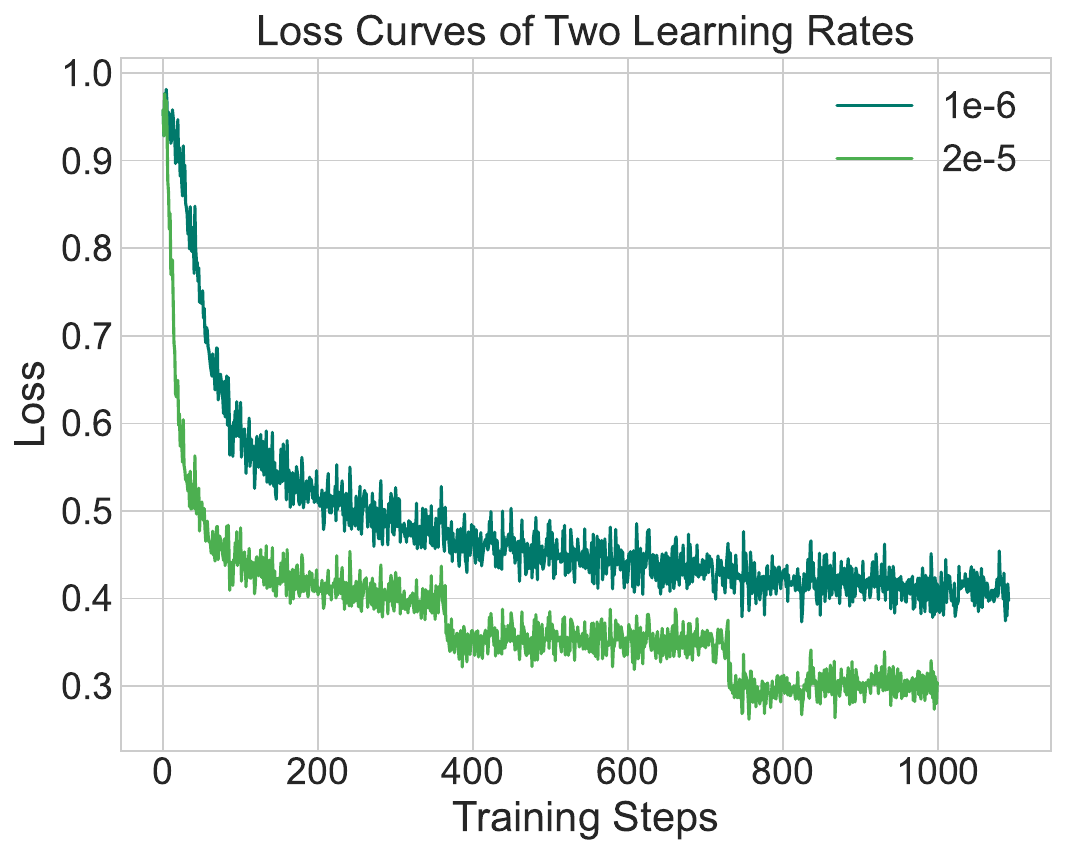}
  \includegraphics[width=\columnwidth]{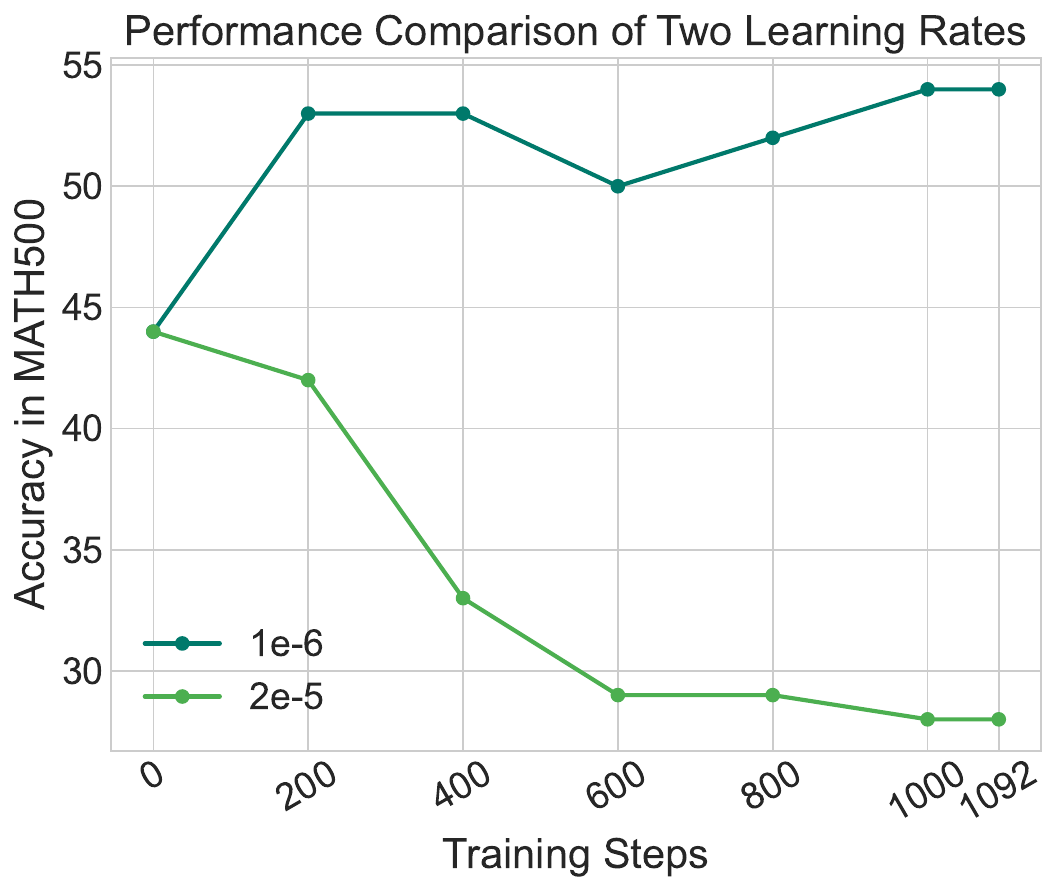}
  \caption {Results of Qwen2.5-3B K\&K distillation. Loss curve shown on \textbf{top} \& performance on MATH500 shown on \textbf{bottom}. Choosing a higher learning rate, $2e-5$ has a chance of overfitting to the K\&K domain rather than learning generalizable reasoning behaviors.}
\label{fig:learning-rate}
\end{figure}

We found that choosing a low learning rate was the best choice for distillation with K\&K data. As shown in Figure \ref{fig:learning-rate}, a lower learning rate leads to a sharp improvement in the performance of the base model in the MATH500 dataset. Since the data is specific to the K\&K domain and might have lower token diversity, it might be possible that a higher learning rate causes the model to overfit quickly to the K\&K task. In comparison, running distillation with s1K data with the learning rate of $2e-5$, keeping the rest of the hyperparameters constant (Table \ref{tab:sft-hyperparams}), did not lead to drastic overfitting. 

To explore another alternative for hyperparameters, we followed the hyperparameter setting from the s1 paper \cite{muennighoff2025s1}, choosing a learning rate of $1e-5$ followed by a cosine decay schedule in contrast to our constant schedule, and training for 5 epochs instead of 3 (Table \ref{tab:sft-hyperparams}). Following these hyperparameters, we distilled on the Qwen2.5-14B model with s1K data. We found that the model still underperformed our K\&K distillation. This new setting led to a performance of $48.3 \pm 1.6$ on MATH500, $31.4 \pm 6.0$ on HumanEval\textsuperscript{+}, and $44.2 \pm 0.5$ on MMLU-Pro. With both old and new hyperparameter settings, distillation with s1K data on the 14B model still leads to a significant performance degradation relative to the base model. A deeper study on why this degradation happens in the 14B model might be necessary. We stick to our hyperparameter settings as we found that to give the best output overall for both distillations, with K\&K and s1K, empirically.



\subsubsection{Warmup with 1,000 K\&K Samples}
As mentioned in Appendix~\ref{distillation-appendix}, we use 4,500 K\&K reasoning traces for distillation. This is a larger number than s1K data which composes of high-quality 1,000 data points. Since the questions for the logic game, K\&K, can be generated arbitrarily, we end up using a high number of samples. Still, we look at the performance of distillation with just 1,000 examples (matching s1K data size) on the Qwen2.5-3B model. We still find strong performance gains throughout the three benchmarks: 51.8 ± 1.4 on MATH, 35.1 ± 1.2  on HumanEval\textsuperscript{+}, and 32.9 ± 0.3 on MMLU-Pro. Although these results are slightly lower than the ones reported in Table~\ref{performance-table}, they are still substantial.

\subsection{Distillation with R1 reasoning traces}
\label{r1-distillation}
To ensure that the observed performance gains were not merely due to potential synergies between the teacher and student models from the Qwen family, we evaluated the effect of using reasoning traces from a teacher model of a different origin. In particular, we considered the possibility that the specific language patterns or token distributions in QwQ’s responses might be especially well-suited to eliciting reasoning behavior in Qwen models. To test this, we employed DeepSeek-R1-Distill-Qwen-32B—a model distilled on reasoning traces generated by DeepSeek R1 \cite{guo2025deepseek}—to produce K\&K traces for warmup, following the procedure described in Section~\ref{warmup}. These traces were then distilled into the Qwen2.5-3B model.

This setup still resulted in consistent improvements across all benchmarks: accuracy on MATH rose to $51.2 \pm 1.6$, HumanEval\textsuperscript{+} to $53.5 \pm 4.3$, and MMLU-Pro to $34.6 \pm 0.5$. These results suggest that the warmup benefits are not limited to teacher-student alignment within the same model family.

\subsection{RLVR Experiment Details}
\label{rlvr-experiment-appendix}

We use the R1 prompt (Table \ref{r1-prompt}) across our training and evaluations \cite{guo2025deepseek}. We modify the prompt slightly based on the task for better instruction following. 
\begin{table}[h]
  \centering
  \begin{tabular}{p{0.9\linewidth}}  
    \toprule
    \textbf{R1 prompt} \\
    \midrule
    A conversation between User and Assistant. The user asks a question, and the Assistant solves it. The assistant first thinks about the reasoning process in mind and then provides the user with the answer. The reasoning process and answer are enclosed within \texttt{\textless think\textgreater} \texttt{\textless /think\textgreater} and \texttt{\textless answer\textgreater} \texttt{\textless /answer\textgreater} tags, respectively, i.e., \texttt{\textless think\textgreater} reasoning process here \texttt{\textless /think\textgreater} \texttt{\textless answer\textgreater} final answer inside \texttt{\textbackslash{}boxed\{\}} tag \texttt{\textless /answer\textgreater}. User: You must put your answer inside \texttt{\textless answer\textgreater} \texttt{\textless /answer\textgreater} tags, i.e., \texttt{\textless answer\textgreater} answer here \texttt{\textless /answer\textgreater}. And your final answer will be extracted automatically by the \texttt{\textbackslash{}boxed\{\}} tag.\\
    \texttt{\{prompt\}} \\
    Assistant: \texttt{\textless think\textgreater} \texttt{\textless /answer\textgreater} \\
    \bottomrule
  \end{tabular}
  \caption{Prompt used across training and evaluations}
  \label{r1-prompt}
\end{table}

For RLVR, we use the hyperparameters shared in Table \ref{tab:additional-hyperparams}.

\begin{table}[htbp!]
  \centering
  \begin{tabular}{ll}
    \toprule
    \textbf{Hyperparameter} & \textbf{Value} \\
    \midrule
    optimizer & AdamW \\
    learning\_rate & 1e-6 \\
    beta & 0 \\
    adam\_beta1 & 0.9 \\
    adam\_beta2 & 0.99 \\
    weight\_decay & 0.1 \\
    warmup\_steps & 25 \\
    Learning Rate Scheduler & constant \\
    bf16 & True \\
    bf16\_full\_eval & True \\
    Batch Size (per device) & 4 \\
    Gradient Accumulation Steps & 5 \\
    Gradient Checkpoint & True \\
    num\_generations & 10 \\
    max\_prompt\_length & 256 \\
    max\_completion\_length & 8192 \\
    max\_grad\_norm & 0.1 \\
    use\_vllm & True \\
    vllm\_max\_model\_len & 8192 \\
    temperature & 0.7 \\
    top-p & 0.95 \\
    GPUs & 6 H100s \\
    \bottomrule
  \end{tabular}
  \caption{RLVR training hyperparameters}
  \label{tab:additional-hyperparams}
\end{table}

\begin{table}[h]
\centering
\resizebox{\columnwidth}{!}{%
\begin{tabular}{lcc}
\toprule
\textbf{Task} & \textbf{Base Model} & \textbf{Warmed-up Model} \\
\midrule
MATH      & 2:00  & 4:00 \\
History   & 3:30  & 3:30 \\
Physics   & 6:30  & 11:00 \\
HumanEval\textsuperscript{+} & 3:30  & 4:00 \\
\bottomrule
\end{tabular}
}
\caption{RLVR Training Times (hh:mm) for Base vs. Warmed-up Models Across Tasks}
\label{tab:compute-resources}
\end{table}

\begin{figure*}[h] 
    \includegraphics[width=0.48\linewidth]{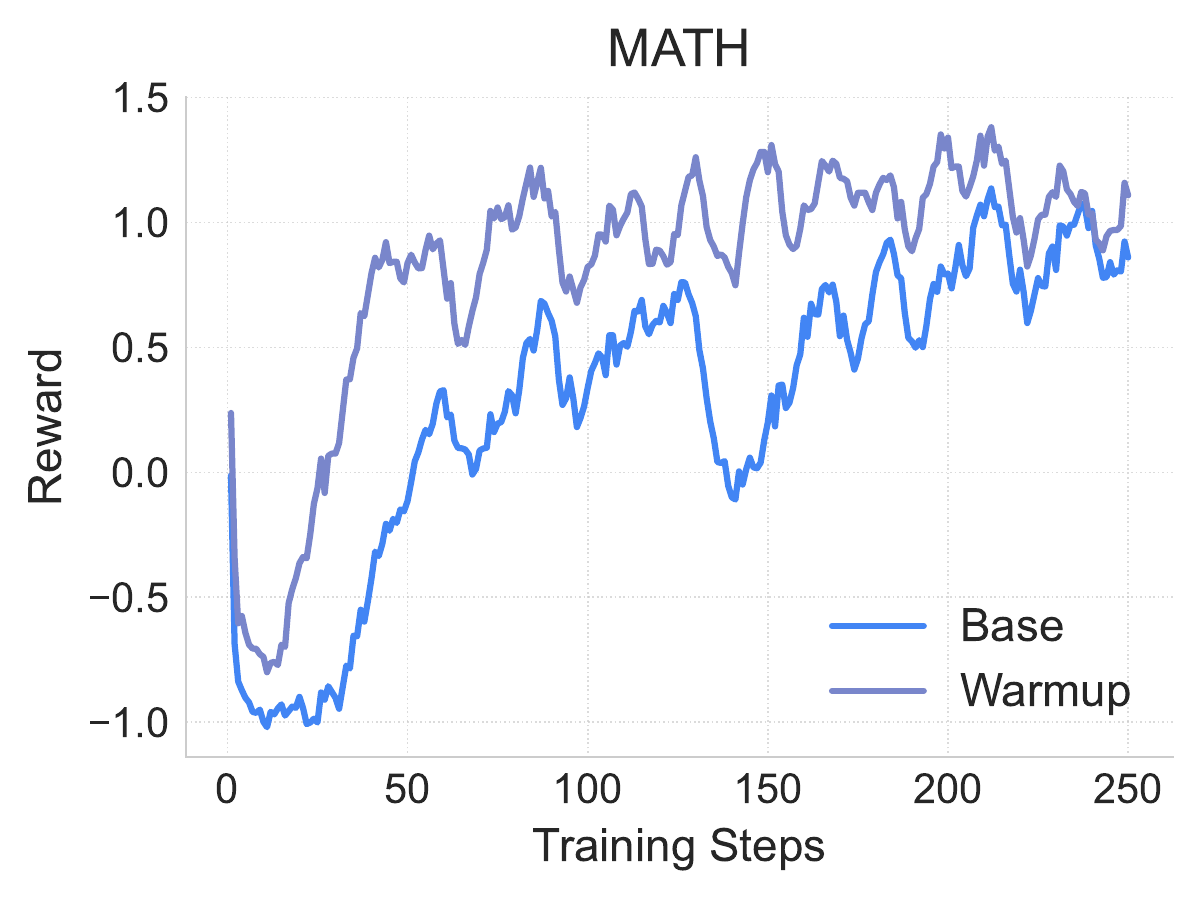} \hfill
    \includegraphics[width=0.48\linewidth]{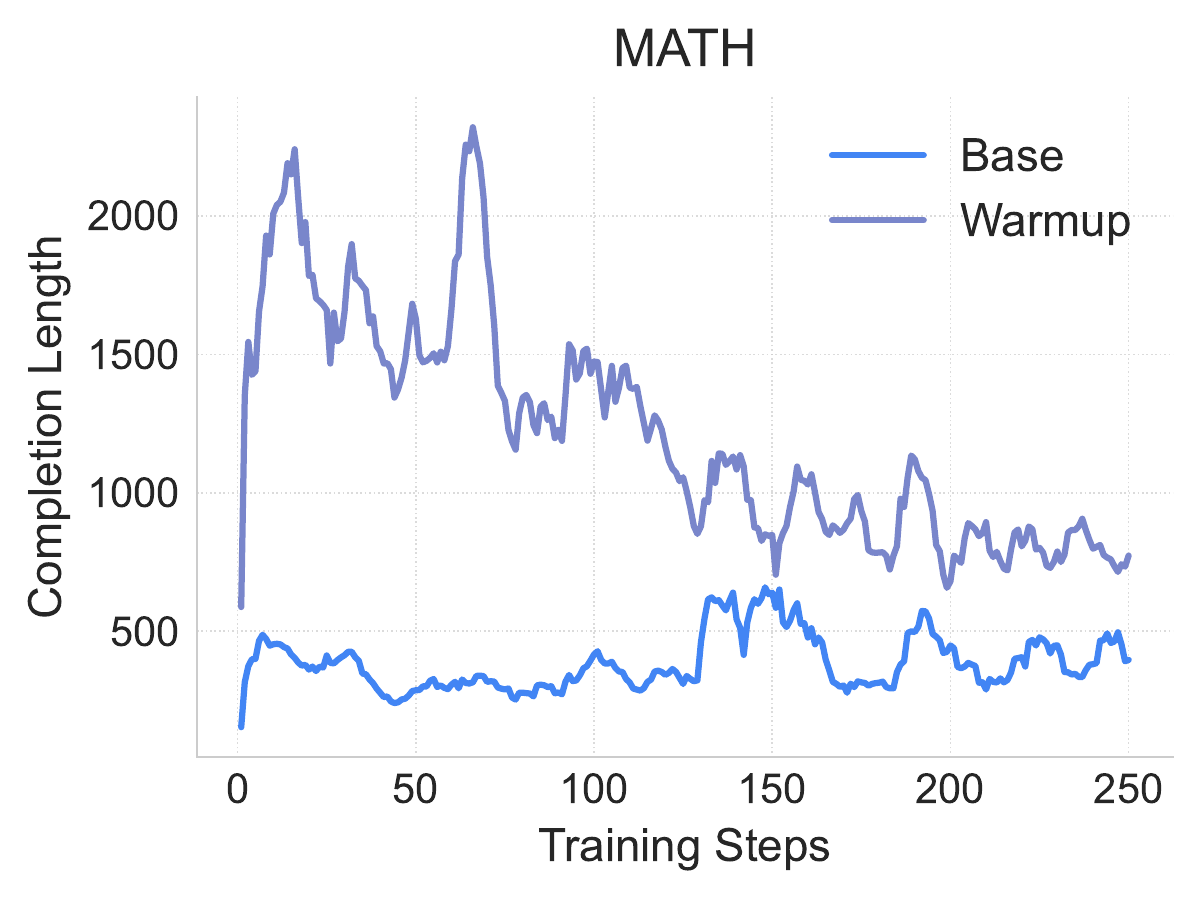}
    \caption{Training curves for Reward \textbf{(left)} and Completion Length \textbf{(right)} for MATH training, smoothed with a moving average (window size 10)}
    \label{fig:math_curves}
\end{figure*}

\begin{figure*}[h] 
    \includegraphics[width=0.48\linewidth]{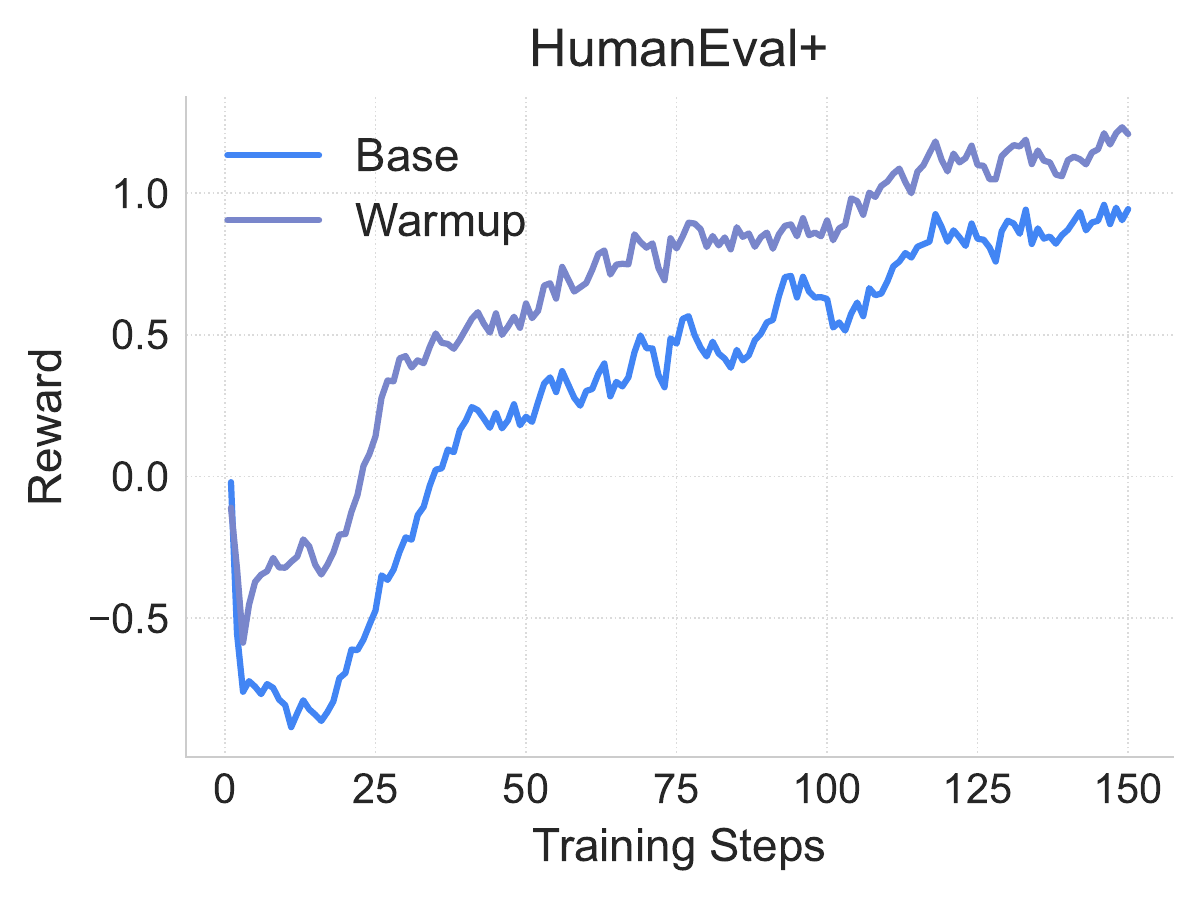} \hfill
    \includegraphics[width=0.48\linewidth]{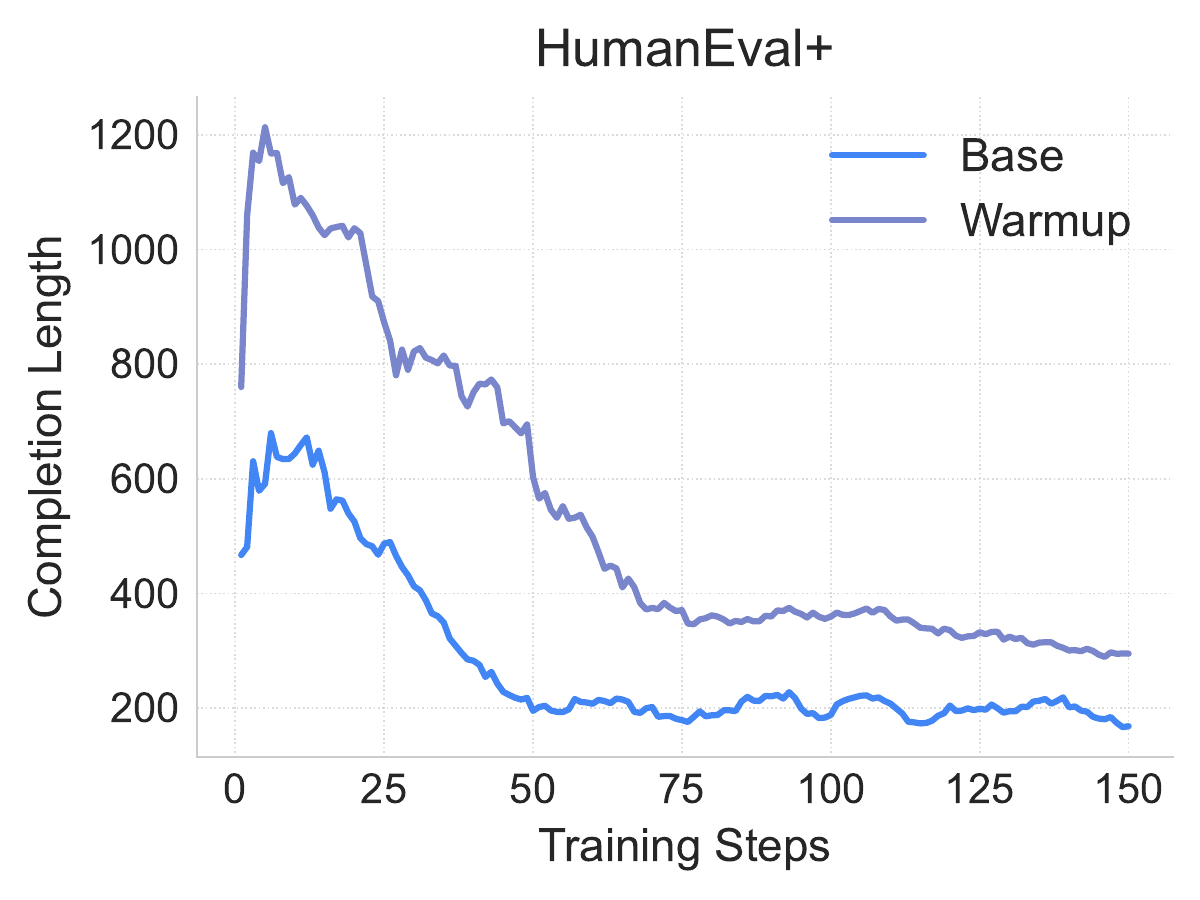}
    \caption{Training curves for Reward \textbf{(left)} and Completion Length \textbf{(right)} for HumanEval\textsuperscript{+} training, smoothed with a moving average (window size 10)}
    \label{fig:humaneval_cruves}
\end{figure*}

Training curves of reward and completion length for MATH (Figure \ref{fig:math_curves}), HumanEval\textsuperscript{+} (Figure \ref{fig:humaneval_cruves}), Physics (Figure \ref{fig:physics_curves}), and History (Figure \ref{fig:history_curves}) are shown below. 

Accuracy and completion lengths (in tokens) for MATH (Table \ref{math-results} and Table \ref{comp-math-results}), HumanEval\textsuperscript{+} (Table \ref{coding-results} and Table \ref{comp-coding-results}), Physics (Table \ref{physics-results} and Table \ref{comp-physics-results}), and History (Table \ref{history-results} and Table \ref{comp-history-results}) are shown below.

\begin{figure*}[h] 
    \includegraphics[width=0.48\linewidth]{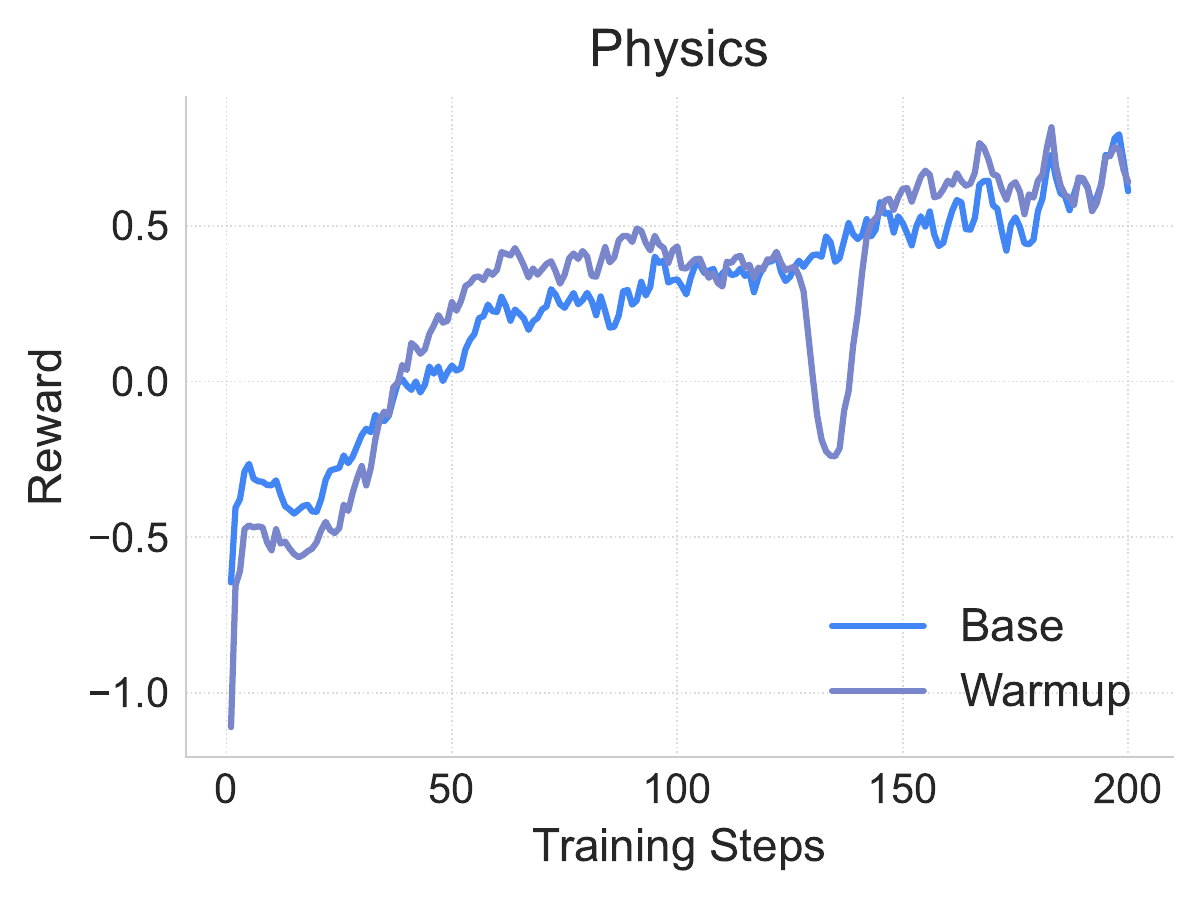} \hfill
    \includegraphics[width=0.48\linewidth]{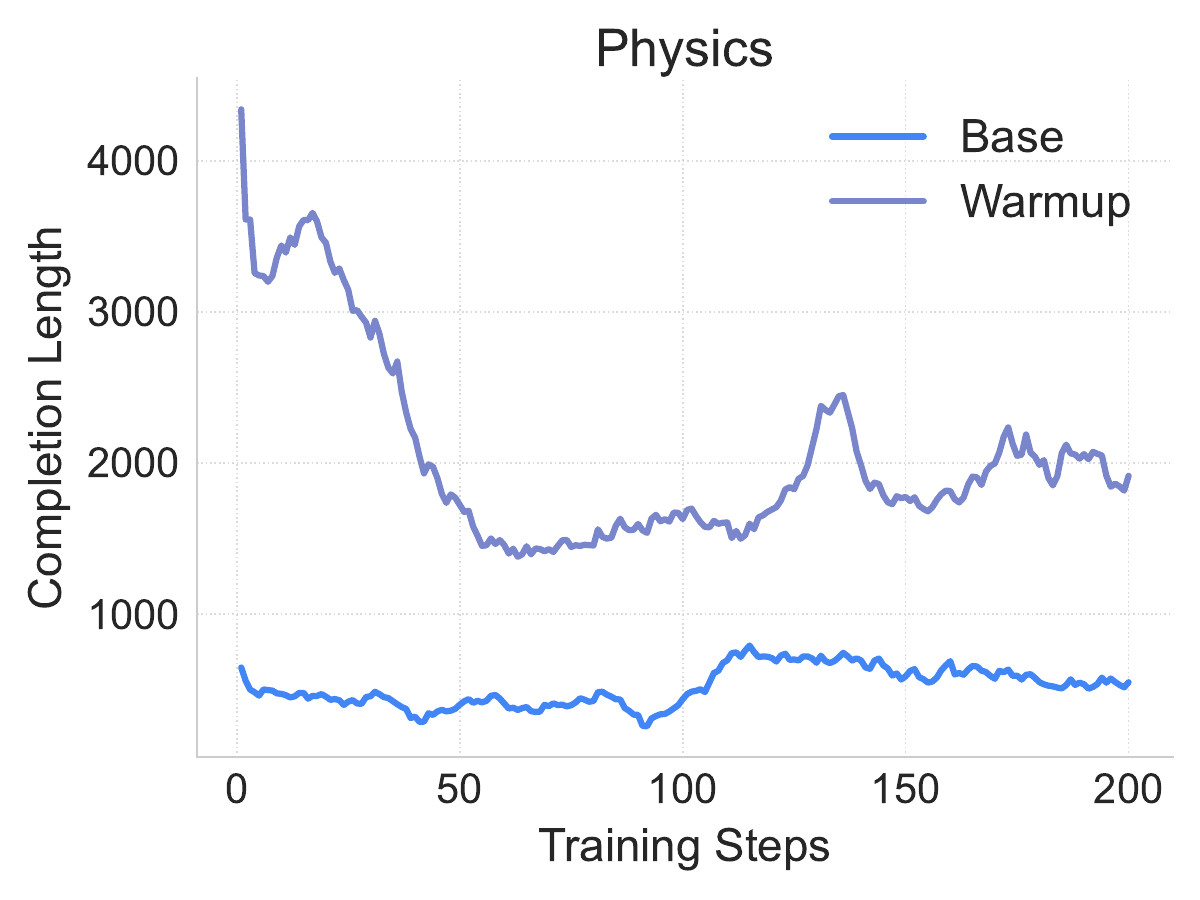}
    \caption{Training curves for Reward \textbf{(left)} and Completion Length \textbf{(right)} for Physics training, smoothed with a moving average (window size 10)}
    \label{fig:physics_curves}
\end{figure*}

\begin{figure*}[h] 
    \includegraphics[width=0.48\linewidth]{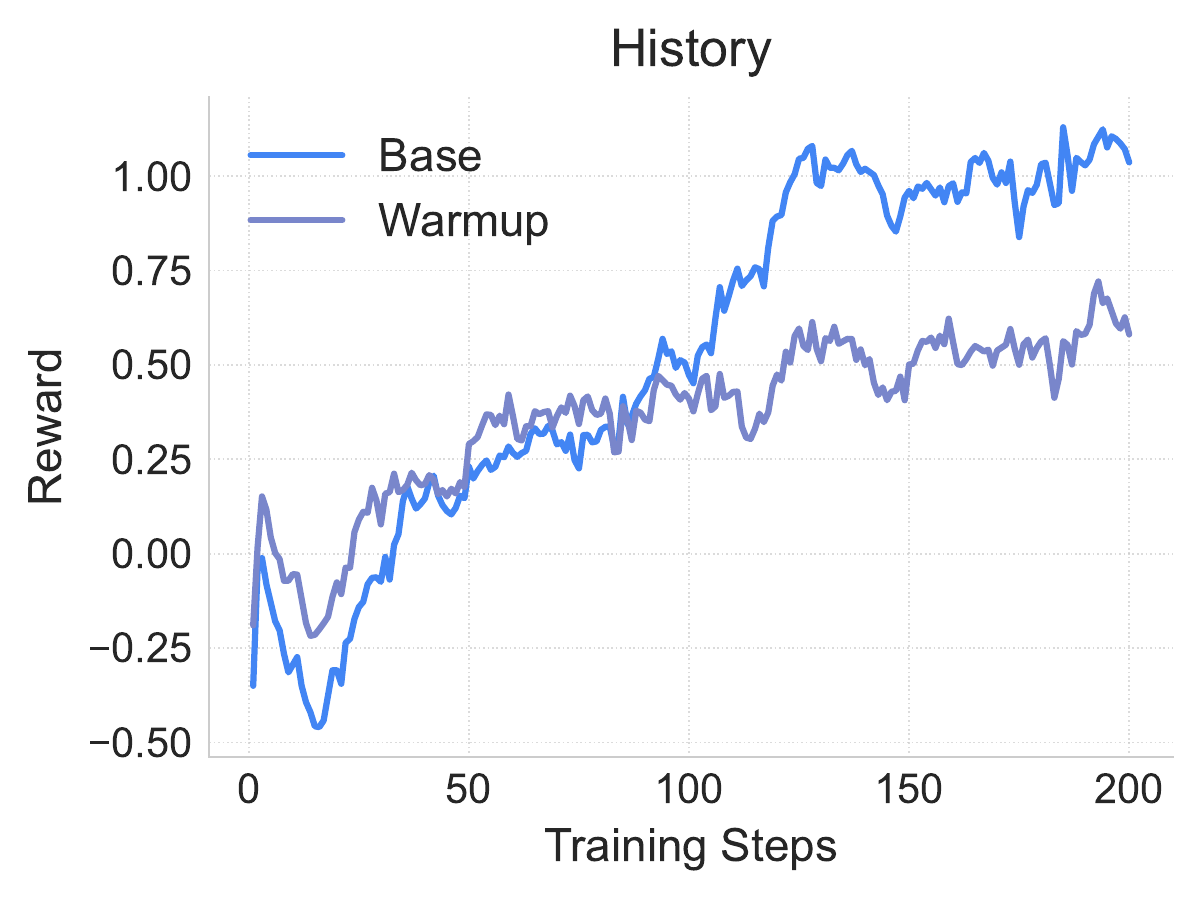} \hfill
    \includegraphics[width=0.48\linewidth]{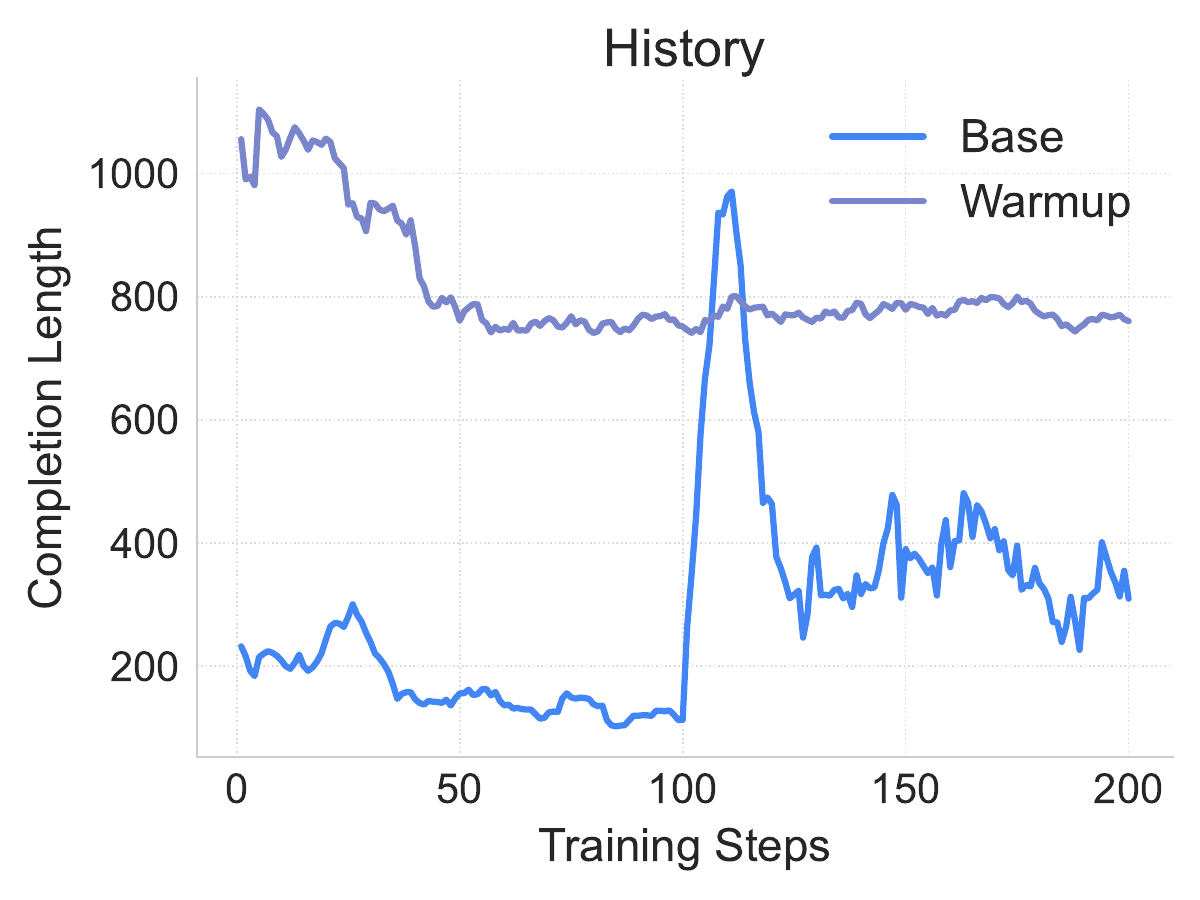}
    \caption{Training curves for Reward \textbf{(left)} and Completion Length \textbf{(right)} for History training, smoothed with a moving average (window size 10)}
    \label{fig:history_curves}
\end{figure*}

\begin{table*}[h]
  \centering
  \begin{tabular}{@{}lccc@{}}
    \toprule
    \textbf{Model} & \textbf{MATH (\%)} & \textbf{HumanEval\textsuperscript{+} (\%)} & \textbf{MMLU-Pro (\%)} \\
    \midrule
    Base Model         & 43.8 ± 0.8 & 32.5 ± 1.2       & 29.2 ± 0.3 \\
    \quad + RL         & 57.8 ± 1.9 &  25.7 ± 4.4              & 28.8 ± 0.2 \\
    \addlinespace
    Warmup             & 54.0 ± 1.4  & 47.8 ± 3.7          & 38.2 ± 0.4  \\
    \quad + RL         & 64.5 ± 0.3 & 31.1 ± 4.8          & 39.2 ± 0.3  \\
    \bottomrule
  \end{tabular}
\caption{Performance comparison of training with MATH on Base vs Warmup Model}
\label{math-results}
\end{table*}

\begin{table*}[h]
  \centering
  \begin{tabular}{@{}lccc@{}}
    \toprule
    \textbf{Model} & \textbf{MATH} & \textbf{HumanEval\textsuperscript{+}} & \textbf{MMLU-Pro} \\
    \midrule
    Base Model         & 696.9 ± 84.8 & 312.6 ± 67.1       & 309.6 ± 13.6 \\
    \quad + RL         & 503.9 ± 19.8 &  307.5 ± 38.5      & 272.2 ± 5.8 \\
    \addlinespace
    Warmup             & 1623.5 ± 19.7   & 1205.9 ± 94.0          &  2203.7 ± 24.5 \\
    \quad + RL         & 1205.8 ± 33.4 & 1235.3 ± 31.6         & 1629.6 ± 7.1  \\
    \bottomrule
  \end{tabular}
    \caption{Completion Length after training with MATH on Base vs Warmup Model}
  \label{comp-math-results}
\end{table*}

\begin{table*}[h]
  \centering
  \begin{tabular}{@{}lccc@{}}
    \toprule
    \textbf{Model} & \textbf{HumanEval\textsuperscript{+} (\%)} & \textbf{MATH (\%)} & \textbf{MMLU-Pro (\%)} \\
    \midrule
    Base Model         & 32.5 ± 1.2      & 43.8 ± 0.8       & 29.2 ± 0.3 \\
    \quad + RL         & 56.8 ± 3.8       & 30.0 ± 0.9    & 27.4 ± 0.9 \\
    \addlinespace
    Warmup             & 47.8 ± 3.7      & 54.0 ± 1.4   & 38.2 ± 0.4  \\
    \quad + RL         & 61.8 ± 3.9       & 52.6 ± 1.5   & 36.2 ± 0.3 \\
    \bottomrule
  \end{tabular}
    \caption{Performance comparison of training with HumanEval\textsuperscript{+} on Base vs Warmup Model}
    \label{coding-results}
\end{table*}

\begin{table*}[h]
  \centering
  \begin{tabular}{@{}lccc@{}}
    \toprule
    \textbf{Model} & \textbf{HumanEval\textsuperscript{+}} & \textbf{MATH} & \textbf{MMLU-Pro} \\
    \midrule
    Base Model         & 312.6 ± 67.1 & 696.9 ± 84.8        & 309.6 ± 13.6 \\
    \quad + RL         & 172.4 ± 2.0 & 328.0 ± 21.3 & 81.3 ± 0.8  \\
    \addlinespace
    Warmup             & 1205.9 ± 94.0 & 1623.5 ± 19.7      &  2203.7 ± 24.5 \\
    \quad + RL         & 351.0 ± 11.6 & 1269.2 ± 29.6 & 1336.4 ± 12.8  \\
    \bottomrule
  \end{tabular}
    \caption{Completion Length after training with HumanEval\textsuperscript{+} on Base vs Warmup Model}
  \label{comp-coding-results}
\end{table*}

\begin{table*}[h]
  \centering
  \begin{tabular}{@{}lcccc@{}}
    \toprule
    \textbf{Model} & \textbf{Physics (\%)} & \textbf{MMLU-Pro (\%)} & \textbf{MATH (\%)} & \textbf{HumanEval\textsuperscript{+} (\%)}  \\
    \midrule
    Base Model         & 24.4 ± 2.6      & 29.2 ± 0.3       & 43.8 ± 0.8 & 32.5 ± 1.2 \\
    \quad + RL         & 33.6 ± 1.1       & 36.0 ± 0.3      & 46.8 ± 0.9 & 38.4 ± 3.0 \\
    \addlinespace
    Warmup             & 34.4 ± 1.2      & 38.2 ± 0.4   & 54.0 ± 1.4  & 47.8 ± 3.7 \\
    \quad + RL         & 41.6 ± 1.6  & 40.8 ± 0.3   & 54.3 ± 2.1 & 48.5 ± 6.1 \\
    \bottomrule
  \end{tabular}
    \caption{Performance comparison of training with Physics on Base vs Warmup Model}
  \label{physics-results}
\end{table*}

\begin{table*}[h]
  \centering
  \begin{tabular}{@{}lcccc@{}}
    \toprule
    \textbf{Model} & \textbf{Physics} & \textbf{MMLU-Pro} & \textbf{MATH} & \textbf{HumanEval\textsuperscript{+}}  \\
    \midrule
    Base Model         & 448.1 ± 37.8 & 309.6 ± 13.6 & 696.9 ± 84.8 & 312.6 ± 67.1 \\
    \quad + RL         & 459.9 ± 30.5 & 274.5 ± 7.7 & 601.1 ± 46.4 & 257.2 ± 42.2 \\
    \addlinespace
    Warmup             & 3607.8 ± 30.5 & 2203.7 ± 24.5 & 1623.5 ± 19.7 & 1205.9 ± 94.0 \\
    \quad + RL         & 1821.3 ± 30.5 & 1348.1 ± 7.6 & 1365.5 ± 45.3 & 1371.2 ± 38.3 \\
    \bottomrule
  \end{tabular}
    \caption{Completion Length after training with Physics on Base vs Warmup Model}
  \label{comp-physics-results}
\end{table*}

\begin{table*}[h]
  \centering
  \begin{tabular}{@{}lcccc@{}}
    \toprule
    \textbf{Model} & \textbf{History (\%)} & \textbf{MMLU-Pro (\%)} & \textbf{MATH (\%)} & \textbf{HumanEval\textsuperscript{+} (\%)}  \\
    \midrule
    Base Model         & 24.9 ± 4.8     & 29.2 ± 0.3       & 43.8 ± 0.8 & 32.5 ± 1.2 \\
    \quad + RL         & 39.7 ± 1.9     & 35.3 ± 0.2       & 35.3 ± 1.0 & 37.1 ± 3.2 \\
    \addlinespace
    Warmup             & 30.0 ± 3.3    & 38.2 ± 0.4   & 54.0 ± 1.4  & 47.8 ± 3.7 \\
    \quad + RL         & 35.7 ± 1.6      & 40.5 ± 0.5   & 50.5 ± 1.8  & 49.8 ± 3.9 \\
    \bottomrule
  \end{tabular}
    \caption{Performance comparison of training with History on Base vs Warmup Model}
  \label{history-results}
\end{table*}

\begin{table*}[h]
  \centering
  \begin{tabular}{@{}lcccc@{}}
    \toprule
    \textbf{Model} & \textbf{History} & \textbf{MMLU-Pro} & \textbf{MATH} & \textbf{HumanEval\textsuperscript{+}}  \\
    \midrule
    Base Model         & 207.2 ± 44.1 & 309.6 ± 13.6 & 696.9 ± 84.8 & 312.6 ± 67.1 \\
    \quad + RL         & 360.6 ± 59.3 & 277.9 ± 10.5 & 498.0 ± 55.8 & 199.5 ± 39.9 \\
    \addlinespace
    Warmup             & 1083.1 ± 129.0 & 2203.7 ± 24.5 & 1623.5 ± 19.7 & 1205.9 ± 94.0 \\
    \quad + RL         & 818.2 ± 5.7 & 1112.1 ± 7.9 & 1117.1 ± 48.3 & 862.8 ± 42.6 \\
    \bottomrule
  \end{tabular}
    \caption{Completion Length after training with History on Base vs Warmup Model}
  \label{comp-history-results}
\end{table*}

\subsection{Compute Costs}

The computational requirements varies across models. The smallest model 1.5B takes around 20 minutes for K\&K distillation with the largest 14B model taking around 1 hour 10 minutes. Since the s1K data only has 1K examples, it takes around 5 minutes for the 1.5B model and around 25 minutes for the 14B model.
We use 6 NVIDIA H100 SXM GPUs (80GB VRAM) from runpod\footnote{\url{https://www.runpod.io/pricing}} coming at a cost of \$$2.99$ per hour per GPU.

We use the same setup of 6 H100 GPUs for RLVR training. For a specific subject, we train both the base and warmed-up model for the same amount of training steps. Duration of training is shown in Table \ref{tab:compute-resources}.

\subsection{History Samples}
\label{history-samples}

After RLVR training the base model on history questions, it learns direct answering as it seems to be the most efficient strategy \cite{wei2022chain,wang2024mmlu}. However, since the warmed-up model is trained on reasoning, it struggles to learn the most efficient strategy as it is biased towards generating more tokens. A sample of both of these cases is shown in Table \ref{tab:history-example-long}.

\onecolumn
\clearpage
\begin{longtable}{p{0.9\linewidth}}
\toprule 
\textbf{Question} 
The ancient city of Cahokia contained more than  \underline{\hspace{1.5cm}} and a population of around  \underline{\hspace{1.5cm}}. 

Options:

\begin{itemize}
    \item[A. ] '40 residential complexes; 40,000'
    \item[B. ] '50 religious monuments; 5,000'
    \item[C. ] '120 earthen mounds; 10,000'
    \item[D. ] '70 public squares; 7,000'
    \item[E. ] '80 ceremonial mounds; 50,000'
    \item[F. ] '200 pyramidal mounds; 200,000'
    \item[G. ] '60 burial mounds; 2,000'
    \item[H. ] '30 agricultural terraces; 30,000'
    \item[I. ] '150 defensive walls; 15,000 '
    \item[J. ] '100 trading posts; 1,000'
\end{itemize}\\
\midrule

\textbf{Correct Answer} (C) \\
\midrule

\textbf{Sample Response after RLVR Training with History questions on base model}\\
The ancient city of Cahokia was home to more than 120 earthen mounds. The population of Cahokia was around 10,000. </think><answer> C </answer>\\
{\color{teal} \textbf{Correct}.}\\
\midrule

\textbf{Sample Response after RLVR Training with History questions on warmed-up model}\\
The question is about the ancient city of Cahokia, which was a significant Mississippian culture settlement in what is now the United States. The goal is to determine the correct answer from the given options regarding the number of residential complexes, religious monuments, earthen mounds, public squares, ceremonial mounds, pyramidal mounds, burial mounds, agricultural terraces, defensive walls, and trading posts. However, the options provided might be a mix of different types of features, and I need to find the pair that corresponds to the most accurate count for the city's size. \\
First, let me recall that Cahokia was a major urban center in the 11th to 13th centuries AD. The city was notable for its large population and impressive architecture. The question asks for the numbers related to the city's residential complexes, religious monuments, earthen mounds, etc., but the options don't list the actual counts of these features. Therefore, I need to look up the historical data related to Cahokia. From historical records, the estimated population of Cahokia is around 15,000-20,000 people. However, the question's options are about specific counts (like the number of residential complexes, religious monuments, etc.), not the population. \\

So, perhaps the options are meant to be about the architectural or ceremonial features that are associated with the city. Let me check each option one by one: A. 40 residential complexes; 40,000 - The population of Cahokia is much smaller, around 15,000, so this seems too high. B. 50 religious monuments; 5,000 - This also seems too high for a city with a population of 15,000. A. 40 residential complexes; 40,000 - The population of Cahokia is much smaller, around 15,000, so this seems too high. B. 50 religious monuments; 5,000 - This also seems too high for a city with a population of 15,000. C. 120 earthen mounds; 10,000 - Again, the population is around 15,000, so this is too high. D. 70 public squares; 7,000 - This is also too high. E. 80 ceremonial mounds; 50,000 - Again, the population is much smaller. F. 200 pyramidal mounds; 200,000 - This is a very high number for a city with around 15,000 inhabitants. G. 60 burial mounds; 2,000 - The population is around 15,000, so this could be possible but let's see other options. H. 30 agricultural terraces; 30,000 - The population is around 15,000, so this is also high. I. 150 defensive walls; 15,000 - The population is around 15,000, so this is possible. J. 100 trading posts; 1,000 - The population is around 15,000, so this is possible. Now, looking at the options, the ones that could be closer to the actual counts of architectural features in Cahokia would be I and J. However, the options don't list the actual numbers of each feature but rather their counts. \\
Since I and J are about defensive walls and trading posts, which are both architectural features, but not necessarily the most accurate count, perhaps I is more reasonable because defensive walls could be a significant part of the city's infrastructure. However, the question might expect a combination of residential complexes, religious monuments, etc., which are not provided. Alternatively, perhaps the question is testing knowledge of the approximate number of ceremonial mounds or public squares in Cahokia. Since the options are about different features, and the population is the only real constraint, I need to find the option that aligns with the city's size. Cahokia had a large number of earthen mounds, which were used for ceremonial purposes. The number of earthen mounds in Cahokia is estimated to be around 120. Let's see if this can be paired with a possible residential complex count. The user's question says ""The ancient city of Cahokia contained more than \underline{\hspace{1.5cm}} residential complexes and a population of around \underline{\hspace{1.5cm}}."" So, the first part is the number of residential complexes, and the second part is the population. Given that the population is around 15,000, the number of residential complexes would likely be a significant number but not too high. Let me consider option G: 60 burial mounds. This is a lower number but not necessarily impossible, but the question suggests that the number of residential complexes might be higher. Since the options are about different features, perhaps I is the best choice here because it's about a tangible architectural feature that could be part of the city's layout, such as defensive walls. Therefore, the best choice among the given options is I: 150 defensive walls; 15,000. This is because the question asks for a combination of ""more than \underline{\hspace{1.5cm}} residential complexes"" and ""a population of around \underline{\hspace{1.5cm}},"" and the defensive walls could be a significant defensive structure in the city, contributing to the city's layout and thus the ""more than"" part. \\
The population is around 15,000, which aligns with the actual historical estimates for Cahokia. </think> <answer> \texttt{\textbackslash{}boxed\{H\}} </answer>\\
{\color{red} \textbf{Incorrect}.}\\
 \\
\bottomrule
\caption{Base model and warmed-up model responses on a History question after RLVR training on History. While the base model gives a very short response without using any Chain of Thought behavior at all, the warmed-up model outputs a long, complex reasoning chain that overcomplicates things. This shows that direct answering without any Chain of Thought seems to be the more favorable strategy for History questions.}
\label{tab:history-example-long} \\
\end{longtable}

\twocolumn

\end{document}